\documentclass[twoside]{article}

\usepackage[accepted]{aistats2025}
\usepackage{microtype}
\usepackage{graphicx}
\usepackage{subfigure}
\usepackage{booktabs} 
\usepackage{comment}
\usepackage{enumitem}
\usepackage{hyperref}
\usepackage{amsmath}
\usepackage{amssymb}
\usepackage{mathtools}
\usepackage{amsthm}
\usepackage{bbm}
\usepackage{dsfont} 
\usepackage{algorithm}
\usepackage{algorithmic}
\usepackage{float}
\usepackage[capitalize,noabbrev]{cleveref}
\usepackage[textsize=tiny]{todonotes}
\theoremstyle{plain}

\theoremstyle{definition}

\theoremstyle{remark}

\DeclareMathOperator*{\argmax}{argmax} 

\newcommand{\yrcite}[1]{\citeyearpar{#1}}

\usepackage[round]{natbib}

\begin{document}

%
\runningtitle{Towards Regulatory-Confirmed Adaptive Clinical Trials}

%
\runningauthor{Omer Noy Klein, Alihan Hüyük, Ron Shamir, Uri Shalit, Mihaela van der Schaar}

\twocolumn[

\aistatstitle{Towards Regulatory-Confirmed Adaptive Clinical Trials: \\ Machine Learning Opportunities and Solutions}

\aistatsauthor{Omer Noy Klein \And Alihan Hüyük \And Ron Shamir}

\aistatsaddress{Tel-Aviv University \And  University of Cambridge \And Tel-Aviv University}

\aistatsauthor{Uri Shalit \And Mihaela van der Schaar}

\aistatsaddress{Technion \\ Tel-Aviv University \And  University of Cambridge}
]

\begin{abstract}
    Randomized Controlled Trials (RCTs) are the gold standard for evaluating the effect of new medical treatments. Treatments must pass stringent regulatory conditions in order to be approved for widespread use, yet even after the regulatory barriers are crossed, real-world challenges might arise: Who should get the treatment? What is its true clinical utility? Are there discrepancies in the treatment effectiveness across diverse and under-served populations? 
    We introduce two new objectives for future clinical trials that integrate regulatory constraints and treatment \emph{policy value} for both the entire population and under-served populations, thus answering some of the questions above in advance. Designed to meet these objectives, we formulate \textsc{Randomize First Augment Next (RFAN)}, a new framework for designing Phase III clinical trials. Our framework consists of a standard randomized component followed by an adaptive one, jointly meant to efficiently and safely acquire and assign patients into treatment arms during the trial. Then, we propose strategies for implementing RFAN based on causal, deep Bayesian active learning. Finally, we empirically evaluate the performance of our framework using synthetic and real-world semi-synthetic datasets.
\end{abstract}

\section{INTRODUCTION}
\label{sec:introduction}

How will a patient respond to a new treatment? For decades, randomized controlled trials (RCTs) have been used for evaluating the effect of new treatments, and are considered the gold standard for clinical evidence of efficacy and safety. Specifically, Phase III (pre-marketing) trials are typically RCTs designed to evaluate the effectiveness of the new treatment compared to a placebo or to a standard treatment, and are the last and crucial step before gaining regulatory approval for wide dissemination. After approval, Phase IV studies (post-marketing surveillance) provide evidence of the treatment's effectiveness and safety on a large scale in the actual target populations and under the actual treatment \emph{policy}\footnote{Policy here means which patients receive the treatment and which do not.}. However, waiting until Phase IV to learn the full real-world implications of a treatment might be too late, as by this point many patients will have been affected.  
A special concern arises from the under-representation in RCTs of populations such as women, ethnic groups, elderly patients, etc., collectively referred to as \textit{under-served} groups. The discrepancy between trial population and the population who ultimately receives the treatment can significantly affect the generalizability of the Phase III trial findings and the effectiveness of care delivery for these populations. A major reason for the existence of these gaps is the fact that currently trials are often designed so as to maximize the chances of obtaining regulatory approval, and recruiting practices that are often deficient in terms of demographic diversity \citep{food2020enhancing,schwartz2023diverse}. To facilitate reading, we provide a glossary of key terms for clinical trials in \cref{tab:glossary}.

As an illustration of these challenges, consider the case of warfarin, a widely used oral anticoagulant \citep{international2009estimation, huynh2017milestone}. 
Despite being used for almost 70 years, warfarin dosing remains a significant challenge for practitioners, primarily due to the drug's narrow therapeutic range and substantial variability in individual responses. An incorrect dose might lead to serious adverse outcomes.
Although several dose-optimization algorithms have been developed and tested in RCTs, these studies were conducted primarily in White and Asian populations
\citep{asiimwe2021warfarin, asiimwe2022ethnic}. Importantly, there are key genetic differences that influence warfarin response, making studies conducted in Whites/Asians less applicable to many other populations. Yet clinical practice guidelines for warfarin dosing have largely been based on these RCTs. Recent years have seen similar pressing challenges, with the emergence of the COVID-19 pandemic \citep{zame2020machine}. For instance, Paxlovid (nirmatrelvir/ritonavir), an antiviral medication for COVID-19, still faces limited distribution due to budget constraints \citep{pepperrell2022barriers}. Given the variability in patient response \citep{najjar2023effectiveness}, identifying who will most benefit from the treatment remains a critical question. Further real-world examples can be found in \cref{tab:rct_examples}.

\begin{figure}[t]
\begin{center}
\centerline{\includegraphics[width=\columnwidth]{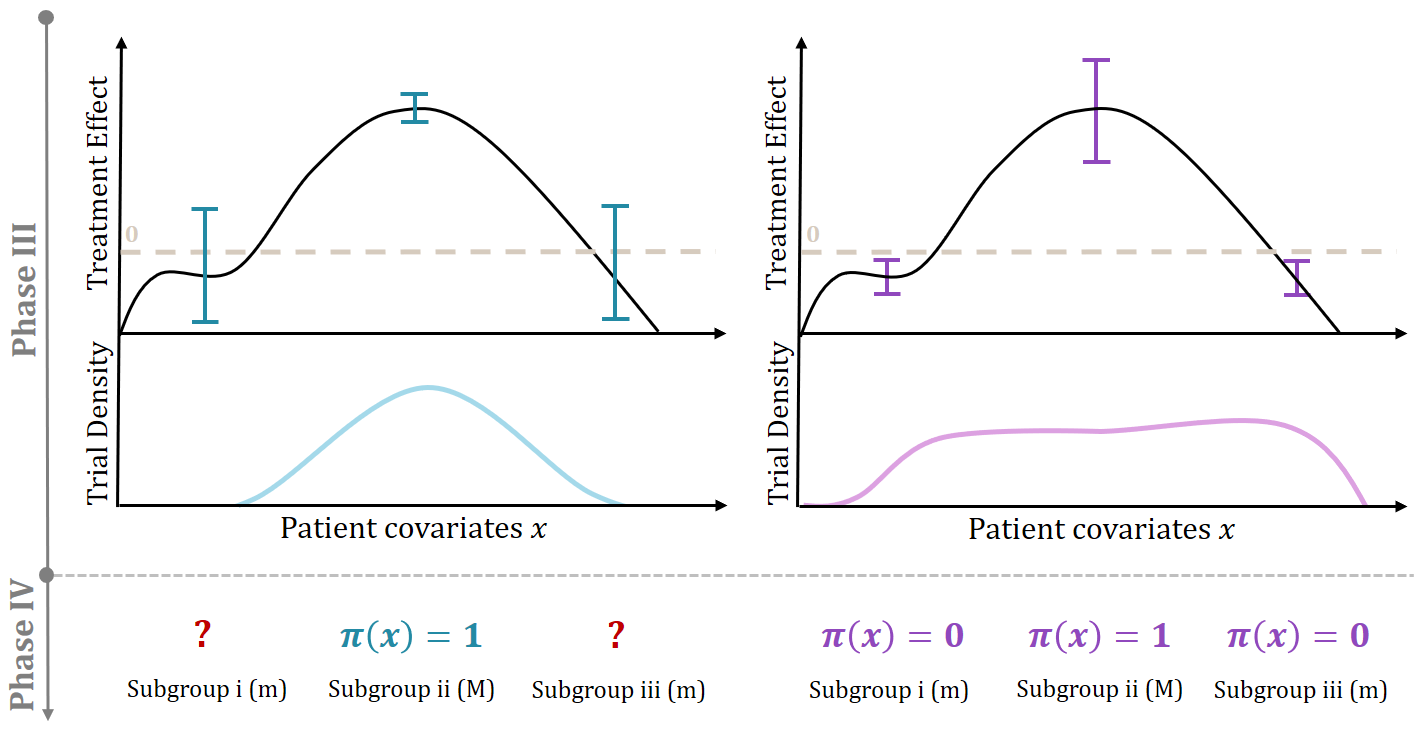}}
\caption{Two Phase III trials with different sample distributions, their impact on the resulting treatment effect estimates (top), and on treatment policy at Phase IV (bottom). Left: Confident treatment effect estimates for the majority (M) subgroup with high uncertainty for the minority groups (m), leading to uncertain treatment policy at Phase IV. Right: While the majority group still can safely benefit from the treatment, better treatment estimates for minority groups are estimated, leading to a beneficial treatment policy.}
\label{fig:illustrated_problem}
\end{center}
\vskip -0.3in
\end{figure}

This paper seeks to formulate a new pre-marketing trial design that bridges some of the gap between the requirements of regulatory bodies and real-world deployment concerns. The fundamental challenges we address in this work are: (1) How can we conduct a representative and fair yet regulatory-confirmed clinical trial? (2) Can we anticipate the actual clinical utility of a new treatment in a way that allows us to deliver a better treatment policy immediately upon deployment? We give an illustration of the challenges in \cref{fig:illustrated_problem}.



We propose to address these challenges by introducing \textsc{Post-Trial Mean Benefit (PTMB)} and \textsc{Post-Trial Fairness (PTF)} as two new objectives for clinical trials, incorporating both the conventional regulatory constraints and the estimated real-world performance of a new treatment with a corresponding policy. Unlike the typical Phase III goal that focuses on regulatory approval, \textit{PTMB also considers the performance of the new treatment policy once deployed in the target population}, for example taking into account the possibility that the new treatment is nominally effective but will have low overall impact. The PTF objective considers the policy value for under-served groups. 

To meet these objectives we propose \textsc{Randomize First Augment Next (RFAN)} as a new framework for designing Phase III clinical trials. The framework consists of two stages: A randomized (\textit{passive}) stage and an augmented (\textit{active}) stage. The randomized stage ensures compliance with regulatory standards. Then, in the augmented stage, the information learned so far is iteratively leveraged for active patient recruitment and treatment assignment. RFAN \textit{adaptively} guides \textit{who to recruit} and \textit{which treatment arm to assign them to}, aiming at prioritizing the exploration of heterogeneous subgroups. Building on ideas from Causal-BALD (Bayesian Active Learning by Disagreement) \citep{jesson2021causal}, we further propose strategies for practical implementation. Finally, we empirically evaluate realizations of our framework on synthetic data and two semi-synthetic real-world datasets of warfarin and COVID-19 patients, showing its potential impact. 

\textbf{Contributions}:
(1) We formulate \textbf{two new objectives}, PTMB and PTF, that we believe align more closely with the goals of post-marketing effectiveness and fairness while incorporating regulatory compliance. (2) We introduce \textbf{\textsc{RFAN}, a new trial design framework} designed to lead to effective and fair treatment policies for the target population while complying with regulatory standards. We present a practical implementation of RFAN trials by adapting novel methods for uncertainty estimation in ML. (3) We empirically evaluate RFAN on synthetic and semi-synthetic real-world datasets and demonstrate its potential impact.


\section{RELATED WORK}
\label{sec:related_work}

\textbf{Adaptive designs} 
The biostatistics community has developed a large body of work on what is known as \textit{adaptive designs} \citep{stallard2014adaptive, thall2021adaptive}. In contrast to the static design of RCTs, an adaptive clinical trial allows dynamic modification of different aspects of its future trajectory, based on intermediate analyses, for better efficiency and ethics. This includes modifications in the trial population, assignment rules, and treatment options as the trial goes on. Notably, major regulatory agencies in Europe and the US have issued detailed guidelines on adaptive designs \citep{pallmann2018adaptive}. Consequently, adaptive designs have received increased attention in recent years. Dominant approaches are \textit{adaptive enrichment} \citep{mehta2009optimizing, magnusson2013group, simon2013adaptive, wang2013adaptive, simon2018using, ondra2019optimized, thall2021adaptive, stallard2023adaptive} and \textit{adaptive signature designs} \citep{freidlin2005adaptive, freidlin2010cross, mi2017enhancement, zhang2017subgroup, bhattacharyya2019adaptive}. However, these designs are limited in their adaptivity and flexibility, as they are restricted to a few pre-specified analysis time points. Adaptive enrichment designs are also limited by the number of pre-specified subgroups they can select between, which is typically two. \cref{tab:adaptive_designs_dimensions} highlights the distinct problem setting of our work compared to adaptive trial designs.

\textbf{Multi-Armed Bandits} 
The setting of a clinical trial is closely related to \textit{pure exploration} problems in the bandit literature. In pure exploration problems the rewards of played arms are not the primary concern, but rather the reward of the singular arm identified at the end \citep{bubeck2009pure, bubeck2011pure, chen2014combinatorial, degenne2019pure}. This approach is studied in Best Arm Identification problems, which aim to identify the arm with the highest mean reward (e.g., \citet{audibert2010best}). In particular, an RCT can be viewed as a purely-exploratory bandit problem aiming to find the best arm for the entire population. Purely-exploratory bandits seek to identify the best arm, but they do not take into account patient-specific information, and thus cannot lead to patient-specific policies including policies that account for how under-served populations respond to treatment. While best-arm identification for contextual bandits has been studied recently, work so far only optimizes for overall reward marginalized over the entire distribution \citep{kato2022semiparametric}, not considering sub-populations or regulatory constraints.
See \cref{appendix:bandit_lit} for details.

\section{PROBLEM SETTING}
\label{sec:problem_setting}

Here we formalize the problem setting and present our newly proposed objectives for clinical trials. 

\textbf{Patients, Treatments \& Outcomes}
At each time point $t\in\{1,\ldots,T\}$ of the experiment (with finite $T$ for practical feasibility), a pool of patients $\tilde{\mathcal{X}}_t\subseteq\mathcal{X}$ with each patient described by covariates $x\in\mathcal{X}$ becomes eligible for enrollment in the experiment (for the time being, we make no assumptions regarding how $\tilde{\mathcal{X}}_t$ evolves). Next, a particular patient $x_t\in\tilde{\mathcal{X}}_t$ among this pool is selected and enrolled in the experiment, and they are assigned to one of two treatment arms $w_t\in\mathcal{W}\allowbreak=\{0,1\}$. Then, an outcome $y_t\in\mathcal{Y}\in\mathbb{R}$ is observed immediately. Following the Neyman-Rubin potential outcomes framework \citep{Rubin1974EstimatingStudies}, we assume that outcomes are determined by two potential outcome distributions $Y^0,Y^1 \in \Delta(\mathcal{Y})^\mathcal{X}$, and we assume causal consistency such that we observe $y_t\sim Y^{w_t}$. We let $\tau(x)=\mathbb{E}[Y^1-Y^0 | x]$ be the conditional average treatment effect (CATE) for patient $x\in\mathcal{X}$, and given a patient distribution $X\in\Delta(\mathcal{X})$, let $\bar{\tau}=\mathbb{E}[Y^1-Y^0]$ be the average treatment effect (ATE). Let $\pi\in\mathcal{W}^\mathcal{X}$ be a treatment policy that maps patients' covariates to treatment assignments. The expected outcome under a particular policy is known as the \textit{policy value}, defined as: $V^{\pi} := \mathbb{E}_{x\sim X,w\sim\pi(x)}[Y^w]$.   
We also denote with $\mathcal{D}_t=\{(x_{t'},w_{t'},y_{t'}\}_{t'=1}^t$ the dataset generated by the entire enrollment and assignment process.

\textbf{Experiment Designs}
The proposed trial design consists of three functions: (i) an acquisition function $\alpha$, (ii) a hypothesis test $\eta$, and (iii) a planning strategy $\rho$. The acquisition function decides which patients are enrolled in the experiment and which treatment arms they are assigned to: $(x_t,w_t)\sim\alpha(\mathcal{D}_t,\tilde{\mathcal{X}}_t)$. The hypothesis test decides whether a treatment effect was detected, i.e. $\eta(\mathcal{D}_T)=1$, or not detected, i.e. $\eta(\mathcal{D}_T)=0$. The hypothesis test is typically strongly regulated and must be passed for the treatment to obtain approval. Finally, the planning strategy describes how to obtain a treatment policy $\pi$ for assigning patients to treatments given the final experimental dataset: $\pi=\rho(\mathcal{D}_T)$. 
We assume that the resulting policy can only be deployed if a valid hypothesis test with type-I error control confirms the efficacy of the new treatment.

\subsection{Objectives}
\label{sec:objectives}

\textbf{Successful Conventional Trial}
The goal of a conventional RCT can be phrased as:
\begin{align*}
    \text{max}_{\alpha \in \mathcal{A_{\text{non-ad}}}}\quad &\mathbb{P}_{\mathcal{D}_T\sim\alpha}\{\eta(\mathcal{D}_T)=1|\bar{\tau}> 0\} \\
    \text{subject to}\quad &\mathbb{P}_{\mathcal{D}_T\sim\alpha}\{\eta(\mathcal{D}_T)=1|\bar{\tau}\leq 0\} \leq \varepsilon,
\end{align*}
where $\varepsilon$ is a predefined error rate (significance level) and $\mathcal{A_{\text{non-ad}}}$ is the set of non-adaptive acquisition functions used in standard RCTs. The degrees of freedom these usually afford are defining apriori inclusion and exclusion criteria (e.g. age bracket, presence or absence of specific comorbidities), and the randomization ratio which is fixed but not always equal for each arm. Conventional clinical trials primarily aim at assessing the treatment's efficacy and safety during the trial, detecting a positive effect, and gaining regulatory approval. 

We now introduce two new trial objectives. Our goal is to develop experiment designs $\alpha$ that attain strong performance on these objectives.

\textbf{Objective 1: Post-Trial Mean Benefit (PTMB)}
We first consider the performance of a treatment policy upon deployment. Denote $V^{\pi}=\mathbb{E}_{x\sim X,w\sim\pi(x)}[Y^w]$ the deployment-time value of a treatment policy $\pi = \rho(\mathcal{D}_T)$, we have for acquisition function $\alpha$:
\begin{align}
    \label{optimal_obj}
    \notag  \text{PTMB}(\alpha) &=  \\ 
    \notag\text{max}_{\rho}\quad &\mathbb{E}_{\mathcal{D}_T\sim\alpha}\left[\eta(\mathcal{D}_T)\cdot V^{\rho(\mathcal{D}_T)} + (1-\eta(\mathcal{D}_T)) V^{\emptyset}\right] \\
    \text{subject to}\quad &\mathbb{P}_{\mathcal{D}_T\sim\alpha}\{\eta(\mathcal{D}_T)=1|\bar{\tau}<0\} \leq \varepsilon.
\end{align}
We denote by $V^{\emptyset}=\mathbb{E}[Y^0]$ the value of the \textit{control policy}, i.e. what we expect the patient outcomes to be had the trial failed and the new treatment is not deployed. Intuitively, for a potentially valuable treatment, PTMB maximizes the estimated overall clinical benefit of the new treatment compared to a standard treatment within the target population ($V^{\pi} - V^{\emptyset}$), had the trial passed ($\eta(\mathcal{D}_T) = 1$). For warfarin, this benefit could mean a reduced risk of adverse events, such as stroke or mortality. 

\textbf{Objective 2: Post-Trial Fairness (PTF)}
We further modify Objective 1 to explicitly capture a notion of fairness. In this work, by fairness we mean that a model does not disproportionately harm a pre-specified group of individuals defined by sensitive attributes such as sex or race. Let $s\in\mathcal{S}$ be sensitive attributes. Denote the conditional policy value $V^{\rho(\mathcal{D}_T)}_s = \mathbb{E}_{w\sim\rho(\mathcal{D}_t)}[Y^w|s]$ and similarly $V^{\emptyset}_s = \mathbb{E}[Y^0|s]$. Here, the goal will be to find experiment designs leading to policies $\pi$ with good max-min fairness \citep{frauen2023fair}:
\begin{align}
    \label{fair_obj}
    \notag  &\text{PTF}(\alpha) =  \\ 
    &\notag \text{max}_{\rho} 
\min\nolimits_{s\in\mathcal{S}}\mathbb{E}_{\mathcal{D}_T\sim\alpha}\left[\eta(\mathcal{D}_T)\cdot
     V^{\rho(\mathcal{D}_T)}_s 
     + (1-\eta(\mathcal{D}_T)) V^{\emptyset}_s\right]\\
     &\text{subject to} \quad \mathbb{P}_{\mathcal{D}_T\sim\alpha}\{\eta(\mathcal{D}_T)=1 \mid \bar{\tau}\leq 0\} \leq \varepsilon.
\end{align}
In words, Objective 2 aims to maximize the worst-case conditional policy value across sub-populations (e.g., across ethnic subgroups in the case of warfarin).

The novelty of our objectives lies in explicitly integrating both trial success and post-trial benefit into a single, forward-looking formulation. In contrast, existing trial designs focus primarily on maximizing trial success (e.g., Type I error control) or, in some cases, consider treatment effect-related measures separately. \cref{tab:adaptive_designs_dimensions} compares the objectives of the existing adaptive trial designs with ours. Globally optimizing \eqref{optimal_obj} and \eqref{fair_obj} is intractable due to the dependence of the policy on the data obtained during the trial. Moreover, directly optimizing would be opaque to regulatory bodies, forming a formidable barrier to adoption. Instead, we next propose a two-stage design striking a balance between regulatory needs and achieving good performance on the objectives. 

\section{RANDOMIZE FIRST AUGMENT NEXT (RFAN)}
\label{sec:rfan}

\begin{figure}[ht]
\begin{center}
\centerline{\includegraphics[width=1\columnwidth]{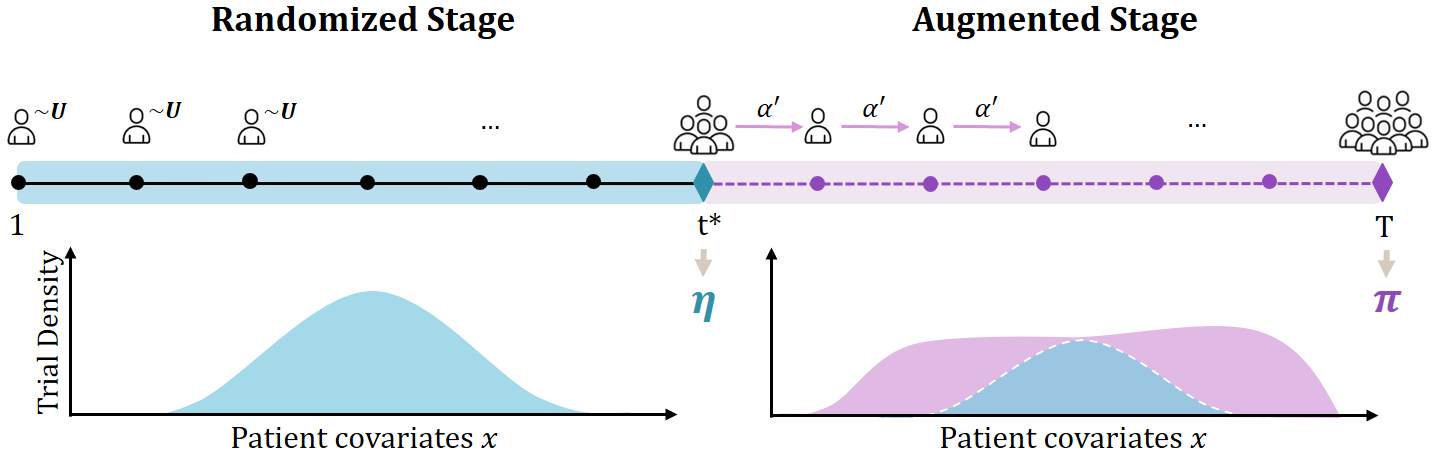}}
\caption{RFAN timeline. $t^*$ and $\alpha'$ are free parameters. The randomized stage establishes a standard regulatory objective ($\eta$), and the augmented stage actively refines a treatment policy $\pi$, jointly addressing our objectives.}
\label{fig:trial_timeline}
\end{center}
\end{figure}

We now introduce \textsc{RFAN}, a framework for designing Phase III clinical trials specifically tailored to deliver robust results on our proposed PTMB and PTF objectives (\cref{sec:objectives}). The trial procedure is outlined in \cref{procedure:cap} and illustrated in \cref{fig:trial_timeline}. Our approach has the following form:
\begin{align*}
    \alpha(\tilde{\mathcal{X}}_t,\mathcal{D}_{t}, \mathcal{M}_{t}) &= \\
    &\begin{cases}
        \text{Uniform}(\tilde{\mathcal{X}}_t\times\mathcal{W}) & t\in\{1,\ldots, t^*\} \\        
        \alpha'(\tilde{\mathcal{X}}_t,\mathcal{D}_{t}, \mathcal{M}_{t}) & t\in\{t^*+1,\ldots, T\},
    \end{cases} \\
     \eta(\mathcal{D}_T) &=\eta(\mathcal{D}_{t^*}) 
\end{align*}
where $\mathcal{M}_t$ is a model fitted to $\mathcal{D}_{t}$ at each time step $t$. This includes fitting a CATE estimator $\hat{\tau}$ to $\mathcal{D}_{t}$ and deriving a treatment policy $\pi_t = \rho(\mathcal{D}_{t})$. RFAN estimates two key quantities of interest: (i) the ATE, estimated in the randomized stage ($t\leq t^*$), addressing regulatory and practical needs, and (ii) the CATE function or the resulting treatment policy for the target population, estimated in the augmented stage ($t^* < t$), depending on the exact setup we use (detailed in \cref{acquisition_selection}). We note that the CATE and policy are functions, not scalar quantities.

RFAN’s randomized stage is necessary to prove an error bound, as codified in regulatory procedures. This can be seen as a restriction of a conventional RCT to the first $t^*$ rounds, with a potential loss of statistical power, to hopefully improve our objectives. The randomized stage retains the strong correctness guarantees of a standard RCT, including an upper bound on its error rate ($\epsilon$). This structured separation ensures that any regulatory-mandated guarantees remain intact, while the augmented stage serves as an empirical extension that optimizes treatment policies under real-world considerations.
For example, running an RFAN trial for warfarin could yield significant benefits by providing improved treatment policy for the population \citep{ma2022warfarin}, and giving practical insights on subpopulations that might benefit more than others.

We next propose possible practical realizations of RFAN through the selection of the acquisition function $\alpha'$ (\cref{acquisition_selection}) and the switching point $t^*$ (\cref{switching_points}).

\begin{algorithm}[htb]
\caption{\textit{Randomize First Augment Next (RFAN)}}
\label{procedure:cap}
\begin{algorithmic}[1]
\REQUIRE Acquisition function $\alpha'$, switching point $t^*$ (predefined or adaptive, see \Cref{switching_points}), acquisition batch size $b$, experimental steps $T$, hypothesis test $\eta$, planning strategy $\rho$
\STATE Initialise: dataset $\mathcal{D}_0 = \emptyset$ 
\STATE \textbf{I) Randomized (Passive) Stage}
\FOR{$t \in\{1, ..., t^*\}$}
    \STATE Acquire a batch of patients $\{x_i\}_{i=1}^b$ from pool $\tilde{\mathcal{X}}_t$ and assign them uniformly at random to treatment arms $\{w_i\}_{i=1}^b$
    \STATE Observe patients' outcomes $\{y_i\}_{i=1}^b$
    \STATE Update: $\mathcal{D}_t = \mathcal{D}_{t-1} \cup \{(x_{i}, w_{i}, y_{i})\}_{i=1}^b$
\ENDFOR \\
\STATE \textbf{II) Augmented (Active) Stage}
\FOR{$t\in\{t^*+1, ..., T\}$}
    \STATE Fit model $\mathcal{M}_{t}$ to $\mathcal{D}_{t-1}$.
    \STATE Estimate $\hat{\tau}_{\mathcal{M}_{t}}(x;\mathcal{D}_{t-1})$ and construct treatment policy $\pi_{t} = \rho(\mathcal{D}_{t-1})$
    \STATE Use $\alpha'(\tilde{\mathcal{X}}_t,\mathcal{D}_{t-1},\mathcal{M}_{t})$ to acquire a batch of patients $\{x'_i\}_{i=1}^b$ and to assign each to treatment arm $\{w'_i\}_{i=1}^b$
     \STATE Observe patients' outcomes $\{y'_i\}_{i=1}^b$ 
    \STATE Update: $\mathcal{D}_t = \mathcal{D}_{t-1} \cup \{(x'_{i}, w'_{i}, y'_{i})\}_{i=1}^b$
    
\ENDFOR
\STATE Conduct hypothesis test $\eta(\mathcal{D}_{t^*})$
\STATE Fit model $\mathcal{M}_{T}$ to $\mathcal{D}_{T}$
\STATE Use $\mathcal{M}_T$ to estimate $\hat{\tau}_{\mathcal{M}_T}(x;\mathcal{D}_T)$ and obtain final treatment policy $\pi_T = \rho(\mathcal{D}_T)$
\end{algorithmic}
\end{algorithm}

\subsection{Selection of Acquisition Function}
\label{acquisition_selection}

We propose using active learning to enroll patients for whom we have the greatest uncertainty into the trial and to assign them to treatment arms. In active learning for treatment effect estimation, a model is trained using existing labeled data that contains patient covariates, assigned treatments, and observed outcomes. Then, an acquisition function is used to assess the informativeness of new samples from data comprised of only covariates and treatments. After acquiring the most informative samples, their outcomes are observed, subsequently leading to model retraining and evaluation. 
We use Bayesian Active Learning by Disagreement (BALD) \citep{houlsby2011bayesian}, which has been shown to be effective when deep learning methods are used. In the Bayesian active learning setup we are given an unlabeled dataset $\mathcal{D}_{\text{pool}} = \{x_i\}_{i=1}^{n_{\text{pool}}}$, a labeled training set $\mathcal{D}_{\text{train}} = \{x_i, y_i\}_{i=1}^{n_{\text{train}}}$, a Bayesian model $\mathcal{M}$ with parameters $\omega \sim p(\omega | D_{\text{train}})$ and predictions $p(y|x,\omega, \mathcal{D}_{\text{train}})$. An oracle provides us with the outcomes \(y\) for any data point in $D_{\text{pool}}$. At each step a batch of data $\{x_i^*\}_{i=1}^b$ is chosen from $D_{\text{pool}}$ using an acquisition function $\alpha$ and the model is retrained. The choice of a Bayesian adaptive approach aligns with extensive existing literature on adaptive clinical trials, as it enables iterative learning under uncertainty and dynamic trial adaptations \citep{berry2010bayesian, williamson2017bayesian, thall2021adaptive, ryan2022bayesian}. Bayesian methods are particularly well-suited for such trials as they model uncertainty of the model parameter estimates (e.g., treatment effect) and are able to iteratively update them as new data becomes available. We note that RFAN is not restricted to the acquisition functions proposed below; alternative acquisition functions could be employed depending on the trial context and objectives.

\textbf{BALD} \citep{houlsby2011bayesian}
uses an acquisition function based on estimating the Mutual Information (MI) between the model predictions and the model parameters  
$I(y ; \omega | x, \mathcal{D}_{\text{train}}) = 
    H(y | x, \mathcal{D}_{\text{train}}) - \mathbb{E}_{p(\omega|\mathcal{D}_{\text{train}})}[H(y | x, \omega, \mathcal{D}_{\text{train}})]$.
The first term is the entropy of the model's prediction, indicating higher values when the model's predictions are uncertain. The second term is an expectation of the entropy of the model prediction over the posterior of the model parameters. In words, BALD scores a data point $x$ based on how well its label $y$ would inform us about the true model parameters. The acquisition function is: $\alpha_{\text{BALD}}(\mathcal{D}_{\text{pool}}, p(\omega | D_{\text{train}})) = \argmax_{\{x_i\}_{i=1}^b} I(y ; \omega | x, \mathcal{D}_{\text{train}})$.

\textbf{Causal-BALD} \cite{jesson2021causal} introduced new acquisition functions for active learning of individual-level causal-treatment effects from high dimensional observational data, based on BALD. In their setting, they assume the pool dataset contains treatment assignments $\mathcal{D}_{\text{pool}} = \{x_i, w_i\}_{i=1}^{n_{\text{pool}}}$, and the labels training set is $\mathcal{D}_{\text{pool}} = \{x_i, w_i, y_i\}_{i=1}^{n_{\text{pool}}}$. \cite{jesson2021causal} have proposed $\alpha_{\mu \text{BALD}}$, among other causally-tailored acquisition functions: $\alpha_{\mu \text{BALD}}(\mathcal{D}_{\text{pool}}, p(\omega | D_{\text{train}})) = \argmax_{\{x_i, w_i\}_{i=1}^b} I(Y^w ; \omega | \{x_i, w_i\}, \mathcal{D}_{\text{train}})$. $I(Y^w ; \omega | \{x_i, w_i\})$ is a measure of the information gain for the model parameters $\omega$ if we obtain a label for the observed potential outcome $Y^w$ given a data point $(x,w)$ and $\mathcal{D}_{\text{train}}$. They show that $I(Y^w ; \omega | \{x_i, w_i\}) \approx \text{Var}_{\omega \sim p(\omega | \mathcal{D}_{\text{train}})}(\hat{\mu}_\omega(x,w))$ where $\hat{\mu}_\omega(x,w)$ is a parametric estimator of $\mathbb{E}[Y| W=w, X=x]$. 


\textbf{Developing Acquisition Functions for Clinical Trials}
Our proposal for $\alpha'$ is built upon $\alpha_{\mu \text{BALD}}$ \citep{jesson2021causal}. Departing from the causal-BALD setup, our approach guides not only \textit{whom to recruit} but also \textit{which treatment arm to assign them to}. In practice, we only have access to a pool of patients with covariates $x$, without treatment assignments. We let $\hat{\tau}_{t}(x)= \hat{\mu}_\omega(x,1)-\hat{\mu}_\omega(x,0)$ be the CATE estimator at time step $t$, based on the model $\mathcal{M}_t$ fit to dataset $\mathcal{D}_{t-1}$. $\hat{\tau}_t(x)$ is then used to inform a treatment policy $\pi_t(x)$. A standard approach for binary treatment is to use the sign of a CATE estimator to construct a policy: $\pi_t(x) = \rho(\mathcal{D}_{t-1}) =  \mathds{1}\{\hat{\tau}(x;\mathcal{D}_{t-1}) > 0\}$.
At each time step $t$, our framework \textit{acquires} a batch of patients $\{x_i\}_{i=1}^b$ and \textit{assigns} each of them into treatment arms $\{w_i\}_{i=1}^b$, aiming to minimize uncertainty in the model parameters to predict the treatment effect. We propose various acquisition functions $\alpha'(\tilde{\mathcal{X}}_t,\mathcal{D}_t)$ for clinical trials, presented in \cref{appendix_tab:acquisition_functions}. Below we present two of the proposed functions:

\textbf{(i) $\alpha_{\mu_{\pi}-Unf}$} acquires patients that maximally reduce uncertainty in model parameters to predict the potential outcome, given the treatment recommended by the current policy $\pi_{t}(x)$. Patients are assigned to treatment arms uniformly at random: $\alpha_{\mu_{\pi}-Unf}(\tilde{\mathcal{X}}_t,\mathcal{D}_t, \mathcal{M}_{t}) = (\{x_i^*\}_{i=1}^b, \{w_i^*\}_{i=1}^b)$, where  
$\{x_i^*\}_{i=1}^b = \argmax_{\{x_i\}_{i=1}^b} I(Y^{\pi_{t}(x)} ; \omega |x, \pi_{t}(x), \mathcal{D}_t)$ and $w_i^* \sim \text{Uniform}(\mathcal{W})$. 

\textbf{(ii) $\alpha_{\text{sign}(\tau)-\pi}$} acquires patients that maximally reduce uncertainty in model parameters predicting the treatment policy sign($\tau$)=$\mathds{1}\{\tau(x) > 0\}$. Patients are assigned treatment according to the policy:
\begin{align*}
            &\alpha_{\text{sign}(\tau)-\pi}(\tilde{\mathcal{X}}_t,\mathcal{D}_t, \mathcal{M}_{t}) = (\{x_i^*\}_{i=1}^b, \{w_i^*\}_{i=1}^b) \\
            &\begin{aligned}
                \notag \{x_i^*\}_{i=1}^b &= \argmax_{\{x_i\}_{i=1}^b} I(\text{sign}(Y^1 - Y^0) ; \omega |x, \mathcal{D}_t) \\
                \notag \{w_i^*\}_{i=1}^b &= \{\pi_t(x_i^*)\}_{i=1}^b.
            \end{aligned}
\end{align*} 
In real-world settings, assigning the correct treatment to a patient is often more clinically important than precisely estimating the potential outcomes. $\alpha_{\text{sign}(\tau)-\pi}$ is built on this principle.

Ideally, our acquisition functions (\cref{appendix_tab:acquisition_functions}) would lead to a better policy value $V^\pi$. We point out that the acquisition functions implicitly account for fairness. First, by actively acquiring patients that maximally reduce uncertainty, they prioritize exploration of underrepresented and/or heterogeneous subpopulations. Second, incorporating the current policy at each time point (e.g., in $\alpha_{\mu_{\pi-Unf}}$) may lead to a more robust policy at the end of the trial. Thus, they can be viewed as approximately optimizing our PTMB and PTF objectives without the regulatory constraints. We demonstrate these phenomena empirically in \cref{sec:experiments}.

\subsection{Selection of Switching Point $t^*$}
\label{switching_points}

For the selection of $t^*$ we propose using early stopping strategies on top of our framework. Specifically, we conduct sequential hypothesis testing at predefined intermediate time points to determine whether to switch to the Augmented Stage or continue in the Randomized Stage. \textit{Alpha spending} functions \citep{demets1994interim}, which are well-established and validated in adaptive clinical trials \citep{o1979multiple, zhang2023systematic}, can be used to responsibly establish adjusted alpha values. We use the \citet{o1979multiple} alpha spending function  $\varepsilon(f) = 2 - 2 \Phi\left(\frac{Z_{\varepsilon/2}}{\sqrt{f}}\right)$, where $\varepsilon$ is the overall significance level, $f \in [0,1]$ is the information fraction at the interim analysis, $Z_{\varepsilon/2}$ is the upper quantile of the standard normal distribution at $\varepsilon/2$ and $\Phi$ is the normal cumulative distribution function. The O'Brien-Fleming boundries are widely used in clinical trials \cite{kumar2016interim, liu2023types}, offering a conservative approach by reducing the risk of falsely concluding efficacy too early. See \cref{appendix:statistical_test_and_es} for more details.

\section{EXPERIMENTS}
\label{sec:experiments}

In this section, we aim to evaluate RFAN and investigate its ability to target our newly established objectives. Evaluation of adaptive designs is a major challenge (\cref{appendix:evaluation}). Since ground truth of treatment effects is unavailable in practice (no access to counterfactual outcomes), a standard approach is using simulated data \citep{friede2012conditional, magnusson2013group, stallard2014adaptive, henning2015closed, rosenblum2016group}, as often done in the literature for treatment effect estimation. We emulate RFAN on synthetic and two semi-synthetic real-world datasets: \textbf{PharmGKB} \citep{limdi2010warfarin}, a dataset of warfarin patients, and \textbf{SIVEP-Gripe} \yrcite{opendatasussrag2021}, a Brazilian COVID-19 hospitalization repository. For details on the experimental setup and implementation\footnote{The code for all experiments can be found at \href{https://github.com/noyomer/rfan-trial}{https://github.com/noyomer/rfan-trial} or at \href{https://github.com/Shamir-Lab/rfan-trial}{https://github.com/Shamir-Lab/rfan-trial}.}, please refer to \cref{appendix:experimental_details} and \cref{sec:appendix_model}.

\begin{table*}[ht]
    \vskip -0.1in
    \caption{Peformance comparison of realizations of RFAN with different acquisition functions (\cref{appendix_tab:acquisition_functions}) against Causal-BALD and RCT (Mean $\pm$ SEM). Error rate ($\epsilon$) is set to 0.05. WC Policy Val.: the worst-case policy value conditioned on sensitive subgroups. ES: Early stopping. Arrows indicate whether higher or lower is better.} 
    \label{tab:performance}
    \begin{center}
    \begin{small}
    (a) Synthetic Data (T=$30$, N=$300$)
    \vskip 0.1in
    \scalebox{0.85}{
    \begin{tabular}{l ccccccc}
        \toprule[1pt]
        \textbf{Design} & \textbf{Policy Val. $\uparrow$} & \textbf{WC Policy Val. $\uparrow$} & \textbf{\% Succ. $\uparrow$} & \textbf{$\sqrt{\epsilon_{PEHE}}$ $\downarrow$} & \textbf{PTMB (Obj. 1) $\uparrow$} & \textbf{PTF (Obj. 2) $\uparrow$} \\ 
        \midrule
         RFAN $\alpha_{\text{sign}(\tau)-\pi}$ ($t^*$=7) & 3.17 $\pm$ 0.02 & 1.23 $\pm$ 0.03 & 0.9 $\pm$ 0.1 & 1.24 $\pm$ 0.12 & 2.95 $\pm$ 0.21 & 1.22 $\pm$ 0.03 \\ 
        RFAN $\alpha_{\mu_{\pi-max}}$ ($t^*$=7) & 3.17 $\pm$ 0.02 & 1.23 $\pm$ 0.03 & 0.9 $\pm$ 0.1 & 0.55 $\pm$ 0.09 & 2.95 $\pm$ 0.21 & 1.22 $\pm$ 0.03 \\ 
        RFAN $\alpha_{\mu_{max}}$ ($t^*$=7) & 3.17 $\pm$ 0.02 & 1.23 $\pm$ 0.03 & 0.9 $\pm$ 0.1 &  0.55 $\pm$ 0.09 & 2.95 $\pm$ 0.21 & 1.22 $\pm$ 0.03 \\ 
        RFAN $\alpha_{\mu_{\pi-Unf}}$ ($t^*$=7) & 3.17 $\pm$ 0.02 & 1.21 $\pm$ 0.04 & 0.9 $\pm$ 0.1 & 0.66 $\pm$ 0.11 & 2.95 $\pm$ 0.21 & 1.2 $\pm$ 0.04 \\ 
        RFAN $\alpha_{\mu_\pi}$ ($t^*$=7) & 3.16 $\pm$ 0.02 & 1.17 $\pm$ 0.03 & 0.9 $\pm$ 0.1 &  1.12 $\pm$ 0.09 & 2.95 $\pm$ 0.21 & 1.16 $\pm$ 0.03 \\ 
        \midrule
        RFAN $\alpha_{\text{sign}\tau-\pi}$ (ES) & 3.17 $\pm$ 0.02 & 1.24 $\pm$ 0.03 & 1.0 $\pm$ 0.0 & 0.87 $\pm$ 0.12 & 3.17 $\pm$ 0.02 & 1.24 $\pm$ 0.03 \\ 
        RFAN $\alpha_{\mu_{\pi}-max}$ (ES) & 3.17 $\pm$ 0.02 & 1.24 $\pm$ 0.03 & 1.0 $\pm$ 0.0 &  0.69 $\pm$ 0.12 & 3.17 $\pm$ 0.02 & 1.24 $\pm$ 0.03 \\ 
        RFAN $\alpha_{\mu-max}$ (ES) & 3.17 $\pm$ 0.02 & 1.24 $\pm$ 0.03 & 1.0 $\pm$ 0.0 & 0.69 $\pm$ 0.12 & 3.17 $\pm$ 0.02 & 1.24 $\pm$ 0.03 \\ 
        RFAN $\alpha_{\mu_{\pi}-Unf}$ (ES) & 3.17 $\pm$ 0.02 & 1.23 $\pm$ 0.04 & 1.0 $\pm$ 0.0 &  0.49 $\pm$ 0.09 & 3.17 $\pm$ 0.02 & 1.23 $\pm$ 0.04 \\ 
        RFAN $\alpha_{\mu_\pi}$ (ES) & 3.17 $\pm$ 0.02 & 1.21 $\pm$ 0.04 & 1.0 $\pm$ 0.0 & 0.82 $\pm$ 0.09 & 3.17 $\pm$ 0.02 & 1.21 $\pm$ 0.04 \\ 
        \midrule
        Causal-BALD & 3.17 $\pm$ 0.02 & 1.22 $\pm$ 0.03 & 0.2 $\pm$ 0.13 & 0.54 $\pm$ 0.06 & 1.43 $\pm$ 0.3 & 1.13 $\pm$ 0.05 \\  
        RCT & 3.09 $\pm$ 0.04 & 0.55 $\pm$ 0.26 & 1.0 $\pm$ 0.0 & 1.13 $\pm$ 0.19 & 3.09 $\pm$ 0.04 & 0.55 $\pm$ 0.26 \\  
        \bottomrule[1pt]
    \end{tabular}}
    \vskip 0.1in
    (b) Warfarin Data (T=$40$, N=$400$)
    \vskip 0.1in
    \scalebox{0.84}{
    \begin{tabular}{lccccccc}
        \toprule[1pt]
        \textbf{Design} & \textbf{Policy Val. $\uparrow$} & \textbf{WC Policy Val. $\uparrow$} & \textbf{\%Succ. $\uparrow$} & \textbf{$\sqrt{\epsilon_{PEHE}}$ $\downarrow$} & \textbf{PTMB (Obj. 1) $\uparrow$} & \textbf{PTF (Obj. 2) $\uparrow$} \\ 
        \midrule
        RFAN $\alpha_{\text{sign}(\tau)-\pi}$ $(t^*=20)$ & \textbf{0.76 $\pm$ 0.01} & \textbf{0.57 $\pm$ 0.03} & 1.0 $\pm$ 0.0 & \textbf{0.82 $\pm$ 0.01} & \textbf{0.76 $\pm$ 0.01} & \textbf{0.57 $\pm$ 0.03} \\ 
        RFAN $\alpha_{\mu_{\pi-max}}$ $(t^*=20)$ & 0.74 $\pm$ 0.01 & 0.52 $\pm$ 0.04 & 1.0 $\pm$ 0.0 & 0.85 $\pm$ 0.02 & 0.74 $\pm$ 0.01 & 0.52 $\pm$ 0.04 \\ 
        RFAN $\alpha_{\mu_{max}}$ $(t^*=20)$ & 0.75 $\pm$ 0.01 & 0.54 $\pm$ 0.06 & 1.0 $\pm$ 0.0 &  \textbf{0.82 $\pm$ 0.03 }& 0.75 $\pm$ 0.01 & 0.54 $\pm$ 0.06 \\ 
        RFAN $\alpha_{\mu_{\pi-Unf}}$ $(t^*=20)$ & 0.75 $\pm$ 0.01 & 0.56 $\pm$ 0.03 & 1.0 $\pm$ 0.0 & \textbf{0.82 $\pm$ 0.01} & 0.75 $\pm$ 0.01 & 0.56 $\pm$ 0.03 \\ 
        RFAN $\alpha_{\mu_\pi}$ $(t^*=20)$ & 0.73 $\pm$ 0.02 & 0.52 $\pm$ 0.04 & 1.0 $\pm$ 0.0 & 0.85 $\pm$ 0.03 & 0.73 $\pm$ 0.02 & 0.52 $\pm$ 0.04 \\ 
        \midrule
        Causal-BALD & 0.75 $\pm$ 0.01 & 0.57 $\pm$ 0.03 & 0.2 $\pm$ 0.2 & 0.83 $\pm$ 0.02 & 0.7 $\pm$ 0.0 & 0.46 $\pm$ 0.01 \\  
        RCT & 0.72 $\pm$ 0.02 & 0.46 $\pm$ 0.03 & 1.0 $\pm$ 0.0 & 0.88 $\pm$ 0.03 & 0.72 $\pm$ 0.02 & 0.46 $\pm$ 0.03 \\ 
        \bottomrule[1pt]
    \end{tabular}}
    \end{small}
    \end{center}
\vskip -0.15in
\end{table*}

\subsection{Datastets}

\textbf{Synthetic Data} 
We modify the synthetic dataset presented in \cite{kallus2019interval, jesson2021causal}:
\begin{align*}
    X \sim & \quad \mathcal{N}(0,1) \\
    Y^w = & \quad (2w - 1)x + (2w - 1) - 2 \sin ((2w - 2)x) + \\
    &\quad 2(1 + 0.5x) + \epsilon, \quad \epsilon \sim \mathcal{N}(0,1).
\end{align*}
We define two sensitive subgroups: $s_1=\{x | x < -1.2 \}$ and $s_2=\{x | x \geq 1.3 \}$, each about $10\%$ of the population. 
Additional details can be found in \cref{appendix:experimental_details}. 

\textbf{Warfarin Data}
We use the PharmGKB dataset \citep{limdi2010warfarin} of 5,700 patients treated with warfarin. The dataset contains clinical and genetic factors that have previously been associated with warfarin, including demographics (age, sex, ethnic group, weight, and smoking status), treatment reason, comorbidities, medications, and genetic factors (presence of genotype variants of CYP2C9 and VKORC1). Importantly, for about $4900$ patients, the dataset includes the true patient-specific stable therapeutic doses of warfarin, which are initially unknown but were adjusted and determined by physicians over a few weeks, until the patient international normalised ratio (INR) is within a target range (the desired levels of anticoagulation). 

We bucket the dosage into two treatment arms: low dosage ($<$ 35 mg/week) and high dosage ($\geq$ 35 mg/week), as in \cite{kallus2022assessing}. We consider binary outcomes that assess the therapeutics' stability and assume for simulation purposes that a warfarin dosage categorized differently from the reported stable one is unstable. Accordingly, for each patient, we set the outcome to $Y=1$ if they were assigned to the arm corresponding to the patient's true optimal dose, and $Y=0$ otherwise. We define race and sex as sensitive attributes. Details can be found in \cref{sec:warfarin_preprocessing}.

\textbf{COVID-19 Data}
We analyze COVID-19 hospitalization data using SIVEP-Gripe (Sistema de Informação de Vigilância Epidemiológica da Gripe) \yrcite{opendatasussrag2021}, a publicly available dataset in the Brazilian Ministry of Health database. Our cohort contains N=11,321 patients who were tested positive for SARS-CoV-2 and hospitalized. We simulate an experiment that assesses the effect of antiviral medications on patient survival. Additional details can be found at \cref{appendix:covid_simulation}.

\subsection{Model}

We use a deep-kernel GP model, which has been shown to perform well on high-dimensional inputs. Deep-kernel GP model uses a deep feature encoder (e.g., Neural network) to transform the inputs and defines a GP kernel over the extracted feature representation to make predictions. We note that, in contrast to \cite{jesson2021causal}, we do not assume access to a validation set, and therefore our experiments maintain active acquisition of a validation set during the trial. Once the trial is done, the model is tuned and retrained using the acquired unseen validation set. We describe model architecture and hyper-parameter space in \cref{sec:appendix_model}.

\subsection{Baselines and Metrics}

While various adaptive clinical trials exist, almost all of them either (i) target different design dimensions to adapt, (ii) operate under different constraints, or (iii) aim to achieve different goals than our trial design. Hence, they are not always suitable benchmarks against RFAN (see \cref{appendix:baselines}). We highlight the distinct problem setting of our work in \cref{tab:adaptive_designs_dimensions}. We evaluate \textsc{RFAN} with our proposed acquisition functions $\alpha'$ (\cref{acquisition_selection} and \cref{appendix_tab:acquisition_functions}) for various values $1<t*<T$. We compare these strategies against: \textbf{(i) RCT:} Uniform patient acquisition and random assignment ($t^*=T$). \textbf{(ii) Causal-BALD:} As a reminder, Causal-BALD (\cite{jesson2021causal}) assumes that $\mathcal{D}_{\text{pool}} = \{x_i, w_i\}_{i=1}^{n_{\text{pool}}}$, i.e., there is no active treatment assignment. We consider $\alpha_{\mu-max}$ (\cref{appendix_tab:acquisition_functions}) as an equivalent version of Causal-BALD with active treatment assignment. To closely emulate active learning only, we set $t^*=1$ for this baseline. Note that $t^*=0$ is infeasible as there is a minimal required number of data points to randomly acquire at the start for model initialization. Although the objective of Causal-BALD is not regulatory constrained, statistical testing is conducted to simulate its use in clinical trials. 

For each dataset, we evaluate the trial designs using different performance metrics: test-set policy value $V^{\rho(\mathcal{D}_T)}$, test-set worst-case policy value over sensitive subgroups $\min\nolimits_{s\in\mathcal{S}} V^{\rho(\mathcal{D}_T)}_s$, success rate $\% \eta(\mathcal{D}_T)=1$,
square root of precision in estimation of heterogeneous effect (PEHE) and our objectives (\cref{sec:objectives}). We discuss measures of adaptive trial designs in \cref{appendix:adaptive_metrics}.

\begin{figure}[t]
\vskip -0.1in
\centering
\subfigure[RCT]{
    \includegraphics[width=0.46\columnwidth]{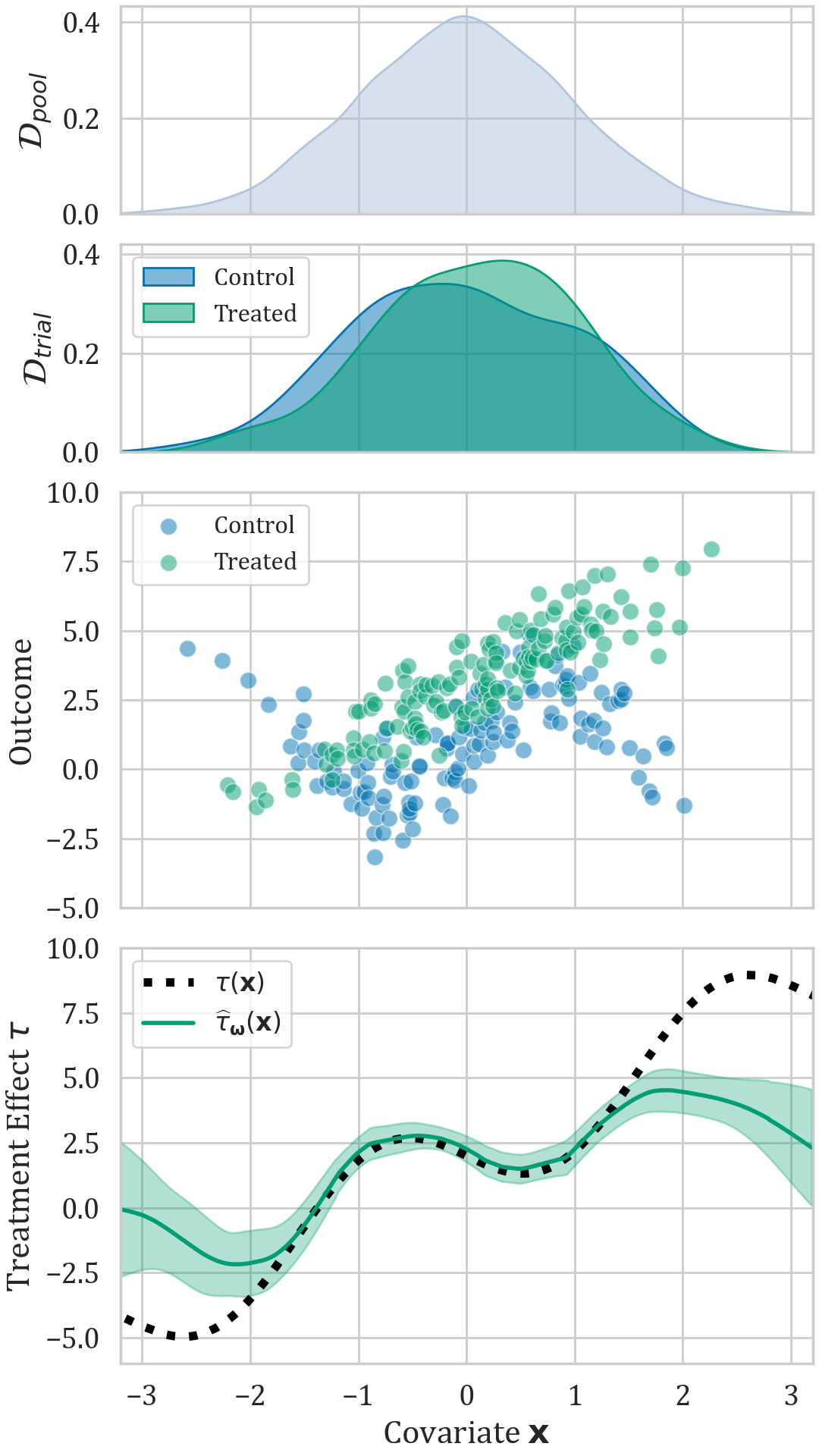}
}
\subfigure[$\alpha_{\mu_{\pi}-Unf}$]{
    \includegraphics[width=0.46\columnwidth]{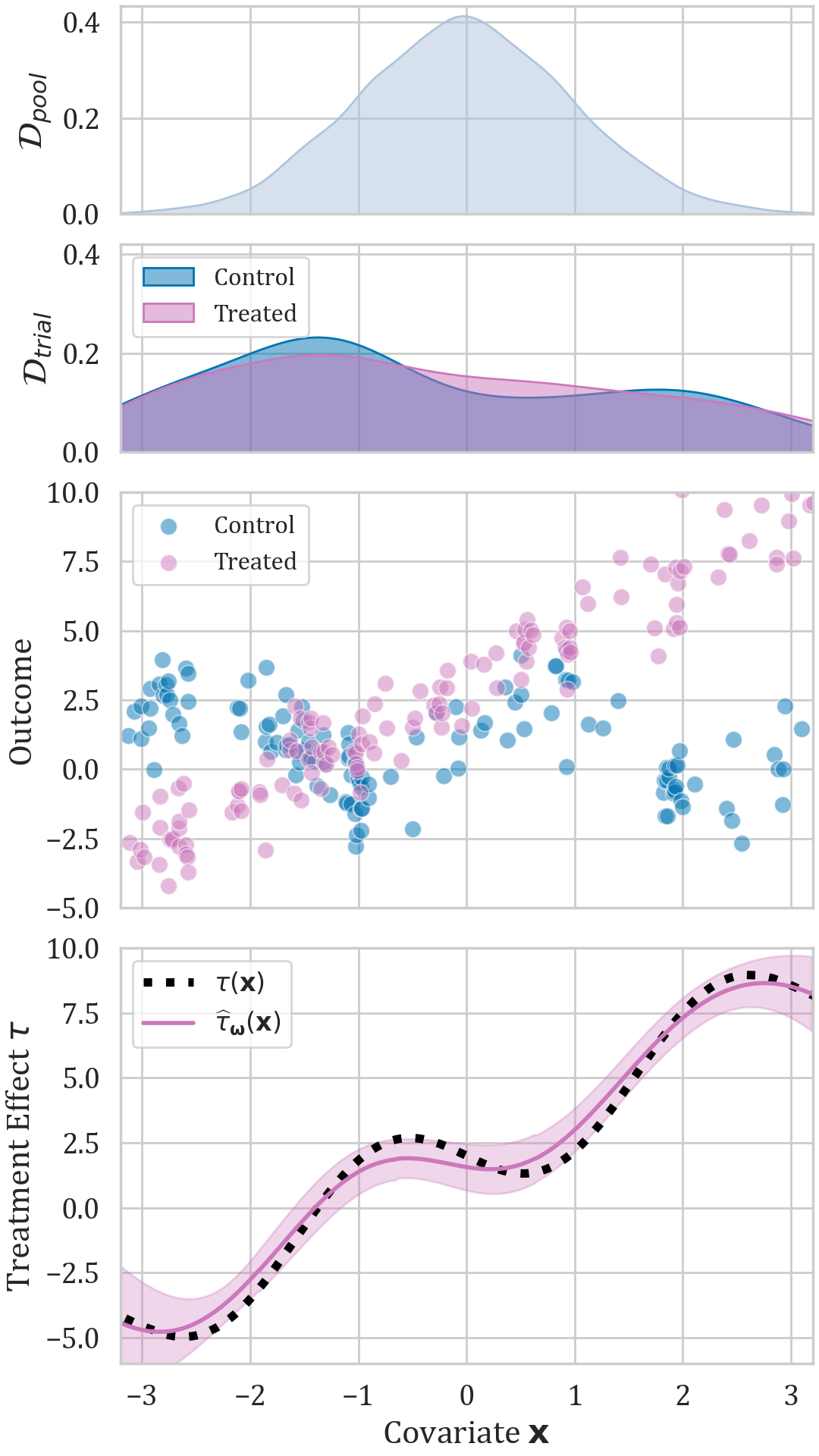}
}
\caption{Distribution of actively acquired synthetic trial population and resulting CATE function.}
\label{fig:data_tau_synthetic}
\end{figure}

\subsection{Experimental Results} 

\cref{fig:data_tau_synthetic} illustrates the effect of each trial design on the distribution of the synthetic trial population (2nd top panel) and on the resulting estimated CATE function. Comparing RCT and RFAN with $\alpha_{\mu_{\pi-Unf}}$, we can see that the population enrolled in the RCT (\cref{fig:data_tau_synthetic}(a)) is an unbiased sample of the pool data (top panel). However, RFAN with $\alpha_{\mu_{\pi-Unf}}$ (\cref{fig:data_tau_synthetic}(b)) demonstrates larger coverage of the target population, and in particular prioritization of enrollment of minority groups. The RFAN CATE estimator is more accurate overall, with better estimation for the sensitive subgroups and slightly worse in the central area.

For the synthetic dataset, we use $T=30$ ($N=300$) and investigate the performance using two predefined switching points $t^*=7$ (\cref{tab:performance}a) and $t^*=15$ (\cref{appendix_tab:performance_300}). The differences between $t^*=7$ and $t^*=15$ illustrate the discussed tradeoff between the success rate and patient benefit. First, we observe that for $t^*=7$, RFAN, using each of the acquisition functions, outperforms RCT in all metrics except success rate and PTMB (Objective 1). Increasing $t^*$ to $15$, the performance of RFAN with any acquisition function outperforms a conventional RCT in all metrics, with a consistent success rate of 100\%. We further investigate the performance of RFAN using early stopping (ES) for selecting $t^*$ (\cref{tab:performance}a), showing the RFAN with any acquisition function outperforms Causal-BALD and RCT, without requiring prior selection of $t^*$. Importantly, while the performance of our framework is comparably good for the Post-Trial Mean Benefit objective, it demonstrates significant improvement in the Post-Trial Fairness (maximizing worst-case policy value conditioned on sensitive subgroups), emphasizing our ability to inform a fair policy without harming, and even improving, the general policy. \cref{tab:performance}b demonstrates similar results for the warfarin dataset using a trial sample size of $N=400$ ($T=40$).

\begin{figure}[t]
\vskip -0.1in
\centering
\subfigure{
    \includegraphics[width=0.29\columnwidth]{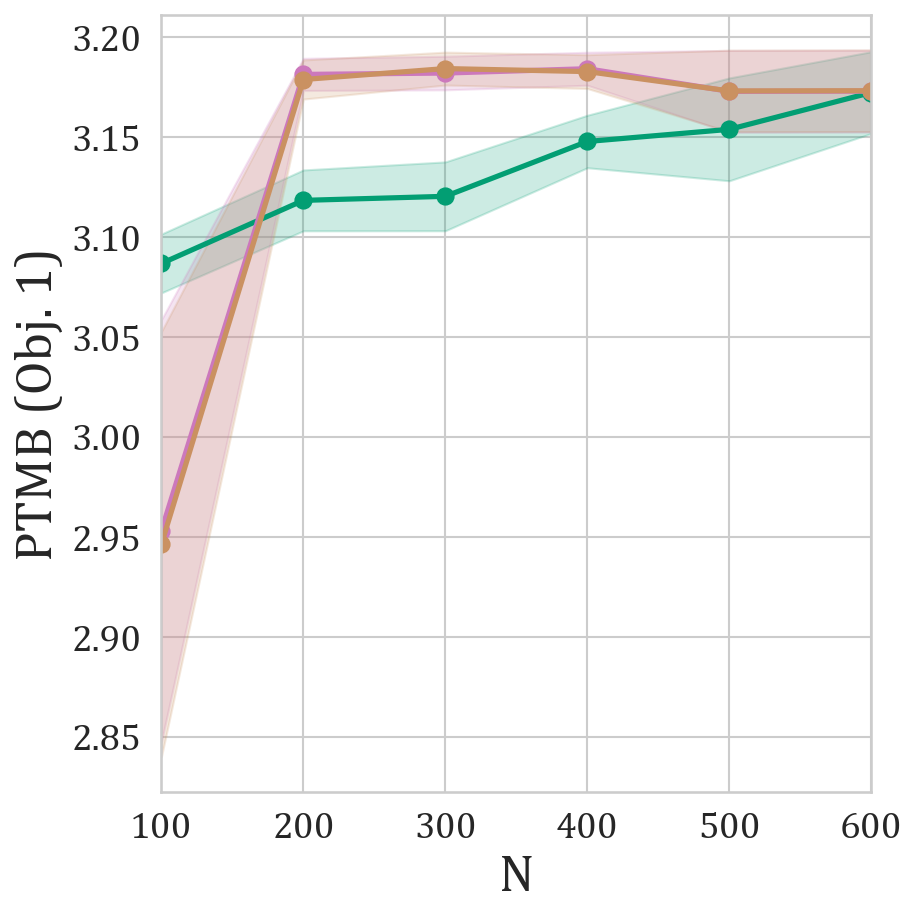}
}
\subfigure{
    \includegraphics[width=0.29\columnwidth]{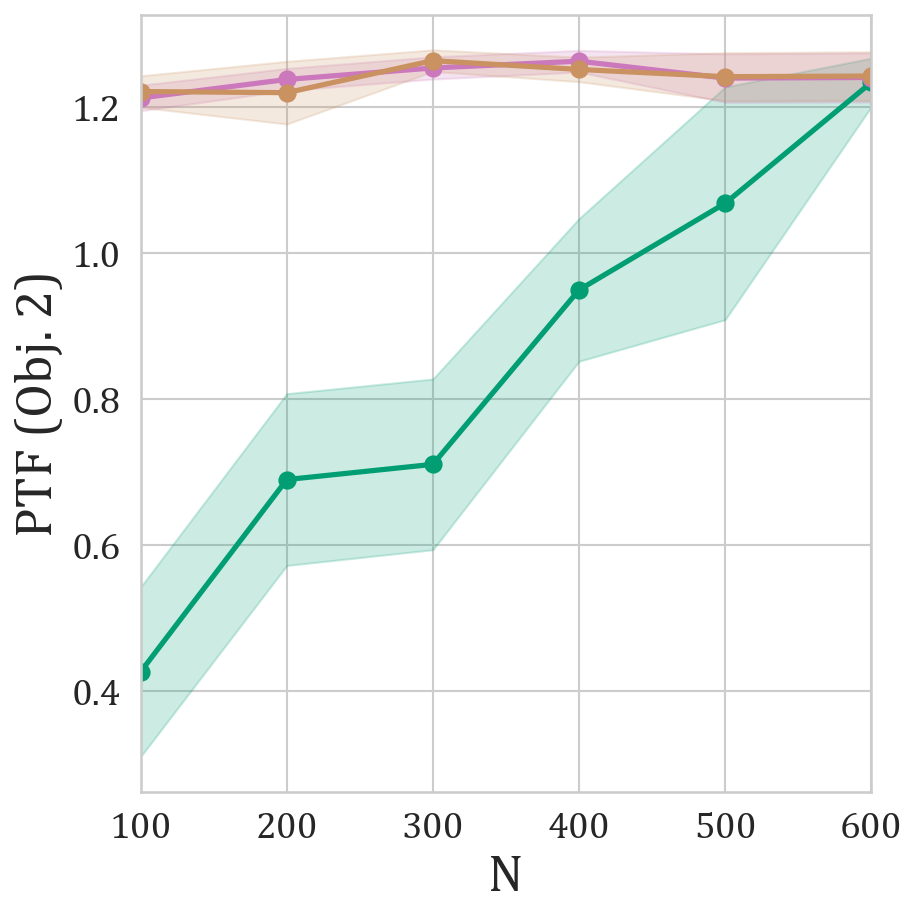}
}
\subfigure{
    \includegraphics[width=0.29\columnwidth]{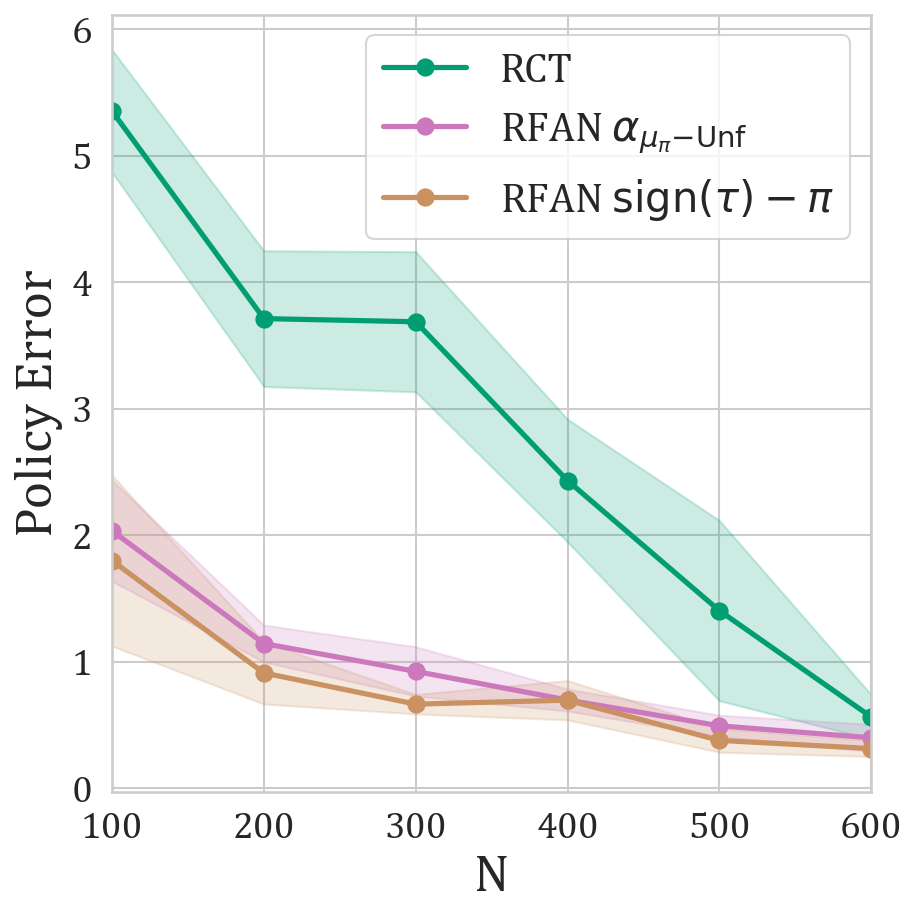}
}
\caption{Performance over varying sample sizes ($N$) on the synthetic data, over $40$ random seeds}
\label{fig:curve_over_N}
\vskip -0.2in
\end{figure}

In an additional experiment, we investigate the performance of RFAN with $\alpha_{\mu_{\pi-Unf}}$ and $\alpha_{\text{sign}(\tau)-\pi}$ against RCT, over varying sample sizes. As can be seen from \cref{fig:curve_over_N}, RFAN decreases the policy error rate for all sample sizes from 100 to 600 patients (reasonable sample sizes of clinical trials). 

\textbf{Supplementary Results} Additional experiments, including those on COVID-19 data, are provided in \cref{appendix:supp_results}.

\section{DISCUSSION}
\label{sec:conclusion}
\textbf{Significance of Main Results} Looking at the synthetic benchmark (\cref{tab:performance}), RFAN with $\alpha_{\text{sign}(\tau)-\pi}$ achieves a two-fold increase in the PTF objective, compared to an RCT, and comparable results in PTMB. The warfarin experiment demonstrates 24\% improvement. As shown in \cref{fig:data_tau_synthetic}, RFAN enables the detection of true treatment effects - both positive and negative - that a conventional RCT might overlook. These differences may lead to significant practical consequences, highlighting RFAN's potential to shape more informed future treatment policies and enhance care for under-served subgroups, without compromising the general population benefit. Note that unlike merely oversampling known underserved groups, our uncertainty-aware acquisition approach enables the recruitment of subgroups with diverse, heterogeneous traits that may not be known in advance or even readily interpretable by the trial designer. This approach aims to address the uncertainty in identifying which patients are likely to benefit from or be harmed by the treatment. 
Importantly, all standard RCT guarantees hold in RFAN's first stage, ensuring regulatory compliance. 

Let us demonstrate RFAN's potential for improving warfarin dosing. An incorrect dose can lead to serious adverse outcomes, including thrombotic events and fatal bleeding. Between 2007 and 2009 warfarin was the leading cause of emergency hospitalizations for adverse drug events in older US adults \citep{budnitz2011emergency}. According to \cite{sombat2023incidence}, the incidence rate of warfarin therapy complications was 4.91 events per 100 person-years. In our context, \textbf{the increase in policy value and reduction in error rate indicate a safer warfarin regimen for sensitive subpopulations}. In turn, this implies fewer hospitalizations and adverse events due to incorrect dosing. This could result in substantial improvements in patient outcomes and significant reductions in healthcare costs. Our results (\cref{fig:curve_over_N}) further show a way for more efficient trials, achieving strong performance with fewer patients, thus preventing unnecessary risk and costs.

RFAN holds significant potential value to key healthcare stakeholders. After treatment approval, \textbf{payers} are particularly interested in \textit{how the treatment benefits specific subpopulations}. RFAN could provide valuable evidence for better resource allocation across diverse populations, enhancing the drug's marketability. In turn, \textbf{caregivers} focus on \textit{how to tailor therapies to individuals} and now could deliver more precise care for heterogeneous populations, which leads \textbf{patients} to benefit from improved clinical outcomes. We discuss ethical and regulatory considerations in \cref{appendix:ethical_considerations} and \cref{appendix:regulatory_considerations}.

\textbf{Conclusion} We investigated how to improve patient welfare and fairness in clinical trials within the real-world regulatory constraints under which clinical trials are held. We examined the existing goals of clinical trials and proposed new objectives that go beyond passing regulatory barriers, bridging the gap between trial success and policy performance in real-world settings. We introduced RFAN, a framework for future designs of clinical trials, designed for achieving our objectives. Although RFAN demonstrates one way to optimize these objectives, we note that there could be other suitable frameworks that require future research. Finally, we present methodological strategies for implementing RFAN based on Causal-BALD (\cite{jesson2021causal}). In synthetic and semi-simulated experiments with real-world patient data, we show that RFAN with uncertainty-aware acquisition functions can lead to significantly better outcomes for under-represented populations without harming the overall benefit of the population and with little or no compromise in the success rate of passing the crucial regulatory barriers.

\subsubsection*{Acknowledgments}
We thank all reviewers for their insightful comments and suggestions. RS was supported by Israel Science Foundation grant No. 2206/22, and the Tel Aviv University Center for AI and Data Science (TAD).  US was
supported by Israeli Science Foundation grant No. 2456/23. ONK was partly supported by a fellowship from the Edmond J. Safra Center for Bioinformatics at Tel Aviv University.

\bibliography{references}




\section*{Checklist}



 \begin{enumerate}

 \item For all models and algorithms presented, check if you include:
 \begin{enumerate}
   \item A clear description of the mathematical setting, assumptions, algorithm, and/or model. [Yes] See \cref{sec:problem_setting} for mathematical setting and \cref{sec:rfan} for a description of our framework.
   \item An analysis of the properties and complexity (time, space, sample size) of any algorithm. [Not Applicable]
   \item (Optional) Anonymized source code, with specification of all dependencies, including external libraries. [Yes]
 \end{enumerate}

 \item For any theoretical claim, check if you include:
 \begin{enumerate}
   \item Statements of the full set of assumptions of all theoretical results. [Not Applicable]
   \item Complete proofs of all theoretical results. [Not Applicable]
   \item Clear explanations of any assumptions. [Not Applicable]     
 \end{enumerate}

 \item For all figures and tables that present empirical results, check if you include:
 \begin{enumerate}
   \item The code, data, and instructions needed to reproduce the main experimental results (either in the supplemental material or as a URL). [Yes] The code is included as part of the supplementary materials and will be publically available upon acceptance. Additionally, see data sharing statements in \cref{appendix:experimental_details}.
   \item All the training details (e.g., data splits, hyperparameters, how they were chosen). [Yes] See \cref{appendix:experimental_details} and \cref{sec:appendix_model}.
          \item A clear definition of the specific measure or statistics and error bars (e.g., with respect to the random seed after running experiments multiple times). [Yes] See \cref{sec:experiments}.
         \item A description of the computing infrastructure used. (e.g., type of GPUs, internal cluster, or cloud provider). [Yes] See \cref{appendix:experimental_details}.
 \end{enumerate}

 \item If you are using existing assets (e.g., code, data, models) or curating/releasing new assets, check if you include:
 \begin{enumerate}
   \item Citations of the creator If your work uses existing assets. [Yes]
   \item The license information of the assets, if applicable. [Not Applicable]
   \item New assets either in the supplemental material or as a URL, if applicable. [Yes] See data sharing statements in \cref{appendix:experimental_details}.
   \item Information about consent from data providers/curators. [Not Applicable]
   \item Discussion of sensible content if applicable, e.g., personally identifiable information or offensive content. [Not Applicable]
 \end{enumerate}

 \item If you used crowdsourcing or conducted research with human subjects, check if you include:
 \begin{enumerate}
   \item The full text of instructions given to participants and screenshots. [Not Applicable]
   \item Descriptions of potential participant risks, with links to Institutional Review Board (IRB) approvals if applicable. [Not Applicable]
   \item The estimated hourly wage paid to participants and the total amount spent on participant compensation. [Not Applicable]
 \end{enumerate}

 \end{enumerate}

\newpage
\appendix
\onecolumn
\aistatstitle{Towards Regulatory-Confirmed Adaptive Clinical Trials: \\ Machine Learning Opportunities and Solutions \\
Supplementary Materials}

\renewcommand{\thefigure}{A.\arabic{figure}}
\renewcommand{\theHfigure}{A\arabic{figure}}

\renewcommand{\thetable}{A.\arabic{table}}
\renewcommand{\theHtable}{A\arabic{table}}
\setcounter{figure}{0}
\setcounter{table}{0}

\begin{table}[H]
    \caption{Glossary of Key Terms in Clinical Trials}
    \label{tab:glossary}
    \begin{center}
    \renewcommand{\arraystretch}{1.5} 
    \setlength{\tabcolsep}{8pt}      
    \begin{small}  
    \begin{tabular}{p{4cm} p{10cm}}
        \toprule[1.5pt]
        \textbf{Term} & \textbf{Description} \\
        \midrule[1.5pt]
        \textbf{Adverse Event} & An unfavorable medical occurrence in a patient after exposure to a treatment (e.g., disease, sign, or symptom). It may or may not be caused by the intervention being studied. \\
        \midrule[0.5pt]
        \textbf{Eligibility Criteria} & A set of rules that determine who can or cannot participate in a clinical study. Eligibility criteria include both \textit{inclusion criteria} (requirements that must be met for a person to participate in the study) and \textit{exclusion criteria} (conditions that prevent a person from participating). \\
        \midrule[0.5pt]
        \textbf{Exclusion Criteria} & A type of eligibility criteria. Specific factors that disqualify individuals from participating in a clinical trial (e.g., children, pregnant women). \\
        \midrule[0.5pt]
        \textbf{Experimental Arm} & A group of participants that receives the treatment being tested in a clinical trial. The group of participants who receive the placebo or the alternative treatment is called the control arm.\\
        \midrule[0.5pt]
        \textbf{Inclusion Criteria} & A type of eligibility criteria. Properties or characteristics that qualify individuals for participation in a clinical trial that define the target population (e.g., age, gender, ethnicity, medical conditions, etc.). \\
        \midrule[0.5pt]
        \textbf{Informed Consent} & The process by which participants are informed about the risks, benefits, and objectives of a study before agreeing to participate. \\
        \midrule[0.5pt]
        \textbf{Observational Study} & A study where participants are observed, without any treatment or intervention assigned by the researchers. \\
        \midrule[0.5pt]
        \textbf{Outcome} & The measured result used to assess the effectiveness of a treatment, such as symptom improvement or adverse effects. \\
        \midrule[0.5pt]
        \textbf{Phase-I trials} & Early-stage clinical trials assessing the safety and dosage of a treatment, typically involving a small number of participants. \\
        \midrule[0.5pt]
        \textbf{Phase-II trials} & Trials focused on evaluating the efficacy and safety of a treatment, often involving a larger group of participants than Phase I. \\
        \midrule[0.5pt]
        \textbf{Phase-III trials} & Trials that are designed to assess the effectiveness of the new treatment over a larger number of participants. They are the final step of research and development before regulatory agencies can approve the new treatment. Phase III trials are nearly always RCTs.\\
        \midrule[0.5pt]
        \textbf{Phase-IV trials} & Post-approval studies conducted to gather additional information on treatment effects in the target population. \\
        \midrule[0.5pt]
        \textbf{Target Population} & The entire population, or group, that a clinical trial is designed to study and draw conclusions about. The target population includes people who meet the eligibility criteria and represent the condition that the intervention/treatment aims to address. \\
        \bottomrule[1.5pt]
    \end{tabular}
    \end{small}
    \end{center}
\end{table}

\newpage
\section{ADDITIONAL LITERATURE REVIEW}

\subsection{Extended Review of RCTs}
\label{appendix: RCTs}

For decades, RCTs have been used for evaluating the effect of new treatments, becoming a gold standard. RCTs often enroll patients from a target population and assign them uniformly at random to treatment groups. Clinical trials are typically classified into four phases.  Phase I and Phase II trials are used to assess safety and efficacy, respectively. Phase III (pre-marketing) trials are nearly always Randomized Controlled Trials (RCTs) and are designed to evaluate the effectiveness of the new treatment compared to a placebo or to standard treatment, and thus gain crucial regulatory approval for wide dissemination. Regulations usually require to control the familywise error rate (FWER), which bounds the probability of a Type 1 error, i.e., the probability of concluding the new treatment is more effective despite not being so. After approval, Phase IV (post-marketing surveillance stage) provides real-world evidence of the treatment’s effectiveness and safety on a large scale in the actual target populations and under the actual treatment assignment policy. While Phase III primarily aims at obtaining regulatory approval, observing and learning real-world implications in Phase IV might be too late.

To illustrate the key challenges involved in the clinical development journey, consider the case of Warfarin. First approved for clinical use in 1954, it is one of the most widely used oral anticoagulant agents worldwide \citep{international2009estimation, huynh2017milestone}.
It is used for the prevention and/or treatment of thromboembolic conditions, such as deep vein thrombosis, rheumatic heart disease, and the prevention of strokes in patients with atrial fibrillation.
Despite being used for almost 70 years, warfarin dosing remains a significant challenge for practitioners, primarily due to the drug's narrow therapeutic range and substantial variability in individual responses. An incorrect dose result in serious adverse outcomes: an insufficient dose can trigger thrombotic events, while an excessive dose might lead to fatal bleeding.
Between 2007 and 2009 warfarin was the leading cause of emergency hospitalizations for adverse drug events in older US adults \citep{budnitz2011emergency}.
Several studies have been conducted to develop dose-optimization algorithms and assess their clinical utility using RCTs. However, most of these studies were conducted in White and Asian populations \citep{asiimwe2021warfarin, asiimwe2022ethnic}. Importantly, there are key genetic differences that influence warfarin response, making studies conducted in Whites/Asians less applicable to many other underrepresented populations. Yet clinical implementation guidelines for warfarin dosing have been deployed based on these RCTs, even though they are mostly applicable to White populations. Recent years have faced similar pressing challenges, with the emergence of the COVID-19 pandemic \citep{zame2020machine}. For instance, Paxlovid (nirmatrelvir/ritonavir), an antiviral medication for COVID-19, still faces limited distribution due to budget constraints \citep{pepperrell2022barriers}. Given the variability in patient response \citep{najjar2023effectiveness}, identifying who will most benefit from the treatment remains a critical question. Further real-world examples can be found in \cref{tab:rct_examples}.

\begin{table}[ht]
    \caption{RCT Examples in the Literature} 
    \label{tab:rct_examples}
    \vskip 0.15in
    \begin{center}
    {\fontsize{8}{10}\selectfont
    \begin{tabular}{p{2cm}|p{1cm}|p{1.8cm}|p{2.2cm}|p{3.2cm}|p{4.4cm}}
        \toprule[1pt]
        \textbf{Study} & \textbf{Size} & \textbf{Treatment} & \textbf{Primary Outcome} & \textbf{Result} & \textbf{Real-World Implications} \\ 
        \midrule[1pt]
        \cite{patel2011rivaroxaban} & 
        14,264 &
        Rivaroxaban vs. dose-adjusted warfarin &
        Stroke or systemic embolism & 
        In patients with atrial fibrillation, rivaroxaban was non-inferior to warfarin & 
        The evidence may be less applicable to underrepresented populations (trial recruited 83\% Whites, 13\% Asians, 1\% Black, and 3\% other) \citep{asiimwe2022ethnic} \\ 
        \midrule
        \cite{polman2006randomized, rudick2006natalizumab}  &
        942; 1,171  &
        Natalizumab vs. placebo & 
        Relapses and disability progression in Multiple Sclerosis (MS) & 
        Natalizumab reduced the risk of the sustained progression of disability and the rate of clinical relapse in patients with relapsing MS & 
        After its approval by the US FDA (2004), Natalizumab was temporarily withdrawn from the market due to cases of progressive multifocal leukoencephalopathy (PML). It was reintroduced in 2006 together with a Global Risk Management Plan \citep{kornek2015update} \\ 
        \midrule
        CAPRIE \yrcite{caprie1996randomised}  &
        19,185  &
        Clopidogrel vs. aspirin & 
        A composite outcome of myocardial infarction, ischemic stroke and vascular death & 
        In patients at risk of ischaemic events, clopidogrel therapy resulted in a relative risk reduction of ischaemic stroke, myocardial infarction, or vascular death, compared with aspirin therapy & 
        Thirteen years after it was approved and marketed, the US FDA issued a warning for poor metabolizers, which are more common in subpopulations other than the majority recruited in the trial (95\% Whites). Such individuals, more common in East Asia and Pacific islands, do not recieve the full benefit of the treatment and therefore remain at risk for heart attack, stroke, and cardiovascular death \citep{wu2015hawaii}. \\ 
        \bottomrule[1pt]
    \end{tabular}}
    \end{center}
\end{table}

\subsection{Extended Review of Bandit Literature}
\label{appendix:bandit_lit}

In recent years, there has been a growing trend in harnessing machine learning to improve the design of clinical trials, with prominent approaches in the multi-armed bandits (MAB) literature. MAB algorithms are particularly useful for clinical trials since they are designed to address the well-studied \textit{exploration-exploitation} trade-off, which can be interpreted as a trade-off between \textit{clinical research} and \textit{clinical practice} in clinical trials. Recent work has focused on the application of MABs to dose-finding \citep{bastani2020online, lee2020contextual, lee2021sdf, aziz2021multi, baek2021fair} and treatment allocation \citep{atan2019sequential, varatharajah2022contextual}. However, while MABs hold the promise of improving patient welfare in clinical trials, there has been a continual hesitance to deploy them in practice \citep{chien2022multi}. The perceived limitations of bandit methods include loss of statistical power, and challenges of statistical analysis, leading to practical barriers to regulatory approval and implementation \citep{villar2015multi, williams2021challenges, robertson2023response}. Some recent work has sought to tackle these challenges from an applied perspective. Villar et al. \citep{villar2015multi, villar2015response, villar2018covariate} proposed a patient allocation strategy for clinical trials, based on the forward-looking Gittins index to improve statistical power. \citet{williamson2017bayesian} introduce a bandit-based design to improve statistical power and reduce treatment effect biases. \citet{williams2021challenges} highlight the difficulties of hypothesis testing with data from bandit models, and propose adjustments to existing statistical tests.

The majority of bandit work on clinical trials primarily focuses on maximizing the rewards of all arms that are ``played'' (e.g., \cite{auer2002finite}), which would have corresponded to maximizing the benefit of enrolled patients \textit{during} the trial (Phase III). Instead, in our design, Phase III is completely exploratory. In that sense, our domain more relates to the concept of \textit{pure exploration} in the bandit literature, where the rewards of played arms are not the primary concern, but rather the reward of the singular arm identified at the end \citep{bubeck2009pure, bubeck2011pure, chen2014combinatorial, degenne2019pure}. This approach is exemplified in Best Arm Identification (BAI) problems, which aim to identify the arm (or the top-K arms) with the highest mean reward (e.g., \cite{audibert2010best}). In particular, an RCT itself can be viewed as a purely-exploratory bandit problem that aims to find the best arm for the entire population. Yet, while the purely-exploratory problems seek to identify the best arm in the trial, they do not explicitly consider the subsequent individual performances of that arm - in our context, the individualized treatment effects in real-world settings (Phase IV). The version of bandits that can deal with patient-specific effects is known s contextual bandits \citep{li2010contextual,chu2011contextual}, where typically the reward is modelled as an inner-product of some unknown arm-specific vector with the instance covariates. An RCT aimed for uncovering a strong policy, without regard to the ongoing reward, could be considered as a best-arm identification contextual bandit problem. So far this has only been studied in the context of learning the best model marginalized over the population, without accounting for sub-populations, fairness, or regulatory constraints \citep{kato2022semiparametric}.




\section{EVALUATION CONSIDERATIONS}
\label{appendix:evaluation}

\subsection{Evaluation of Clinical Trials}

Evaluation of new designs of clinical trials is a major challenge, even in the biostatistics literature. Conducting a clinical trial that evaluates our new framework is infeasible in the scope of our paper, due to complex ethical, legal, and financial considerations. Ethically, ensuring the well-being and informed consent of participants is fundamental, requiring careful planning and approval from ethical review boards. Legally, compliance with strict regulations and guidelines, often varying across jurisdictions, poses a significant barrier. Financially, the substantial costs involved in clinical trials, from patient recruitment to data analysis, are beyond our current resources. Beyond that, one of the challenges in our case is that our objectives are defined not only on the trial's performance but also based on Phase IV performance (i.e., policy value after deployment), had the trial been passed. This requires long-term control and monitoring of newly treated patients in the general public population. Given these complexities, a prospective validation of our framework is beyond the focus of this paper. Instead, simulated data are generally used to evaluate newly proposed clinical trial designs \citep{atan2019sequential, huyuk2022make, curth2023adaptive}, even in the biostatistics literature \citep{friede2012conditional, magnusson2013group, stallard2014adaptive, henning2015closed, rosenblum2016group}. We therefore evaluate our experimental designs on synthetic data, which enables verification with \textit{known} ground truth. Furthermore, we evaluate our framework on \textbf{PharmGKB} \citep{limdi2010warfarin}, a dataset of patients treated with warfarin, and \textbf{SIVEP-Gripe} \yrcite{opendatasussrag2021}, a Brazilian COVID-19 hospitalization repository, for emulating real-world settings. Future work could explore additional evaluation approaches under these limiting considerations.

\subsection{Trial Baselines}
\label{appendix:baselines}

While numerous approaches exist for running adaptive clinical trials, almost all of them either (i) target different design dimensions to adapt, (ii) operate under different constraints, or (iii) aim to achieve different goals than our trial design. Hence, they are not always suitable benchmarks against RFAN. 

For instance, some common types of adaptive trial designs include \citep{chow2008adaptive, chow2011adaptive, thorlund2018key}: (i) adaptive randomization design, where the ratio of patient in each treatment arms is updated to increase power (e.g., \cite{lachin1988statistical}), (ii) group sequential design, which allows for early stopping due to futility (e.g., \cite{lehmacher1999adaptive}), (iii) sample size re-estimation designs, where the sample size is updated with interim analysis to maintain trial power (e.g., \cite{chuang2006sample}), (iv) drop-the-losers design, where ineffective treatment arms are dropped mid-trial (e.g., \cite{sampson2005drop}), (v) adaptive enrichment/signature design, where both the inclusion criteria and the target population of the final analysis can be switched mid-trial (e.g., \cite{simon2013adaptive}), and (vi) adaptive seamless designs, where data from multiple trials are combined in one analysis (e.g., \cite{maca2006adaptive}).

We have summarized the adapted design dimensions and the fixed design dimensions in these trial designs as well as their objectives in \cref{tab:adaptive_designs_dimensions} to highlight the distinct problem setting of our work. Adaptive enrichment/signature designs are the most closely related to RFAN as they also adapt the inclusion criteria. However, in an enrichment/signature design, this is done to maximize the success probability of the trial, often by adapting the target population of the final analysis along with the inclusion criteria (i.e. by changing the null hypothesis that is tested at the end of the trial). In our work, the null hypothesis is treated as a fixed design dimension.

\subsection{Metrics for Evaluating Adaptive Clinical Trials}
\label{appendix:adaptive_metrics}
Adaptive trial designs have been extensively studied for decades, yet there is no widely accepted measure for comparing their performance and this remains an open research question. Adaptive trials are often evaluated based on the probability of trial success, i.e., whether a valid hypothesis test with type-I error control confirms the efficacy of the new treatment (denoted as $\eta(\mathcal{D}_T)=1$ in our paper). In some cases, depending on the design context, other metrics are used, such as measures of trial resources needed (e.g., required sample size \citep{zhang2024reporting}) and measures of ethical considerations (e.g., quantifying patient exposure to inferior treatment \citep{korn2018interim}). However, to our knowledge, there is no gold standard besides statistical guarantees such as power and error rate. In particular, no measure combines both regulatory requirements and the predicted patient benefit. Our primary contribution to this challenge is the introduction of two innovative objectives for future clinical trials – PTMB and PTF – which we view as steps towards closing the discrepancy between clinical trials and the way treatments are deployed in practice.

\begin{table}[h]
    \caption{Adaptive trial designs comparison}
    \label{tab:adaptive_designs_dimensions}
    \vskip 0.1in
    \begin{center}
    \renewcommand{\arraystretch}{1.5} 
    \setlength{\tabcolsep}{8pt}      
    \begin{small}  
    \begin{tabular}{p{3.5cm} p{3.5cm} p{4cm} p{3cm}}
        \toprule[1.5pt]
        \textbf{Adaptive Trial Design} & \textbf{Adapted Design Dimensions} & \textbf{Fixed Design Dimensions} & \textbf{Objective} \\
        \midrule[1.5pt]
        Adaptive Randomization Design & Distribution of patients across treatments & Error rate, Sample size, Target population, Inclusion criteria & Maximizing power \\
        \toprule
        Group Sequential Design & Final sample size \newline($<$ Target sample size) & Error rate, Target population, Inclusion criteria & Minimizing sample size while maintaining power \\
        \toprule
        Sample Size \newline Re-estimation Design & Target Sample Size & Error rate, \newline Target population,\newline Inclusion criteria & Minimizing sample size while maintaining power \\
        \toprule
        Drop-the-Losers Design & Set of treatments & Error rate, Sample size, \newline Target population,\newline Inclusion criteria & Maximizing sensitivity in identifying effective treatments \\
        \toprule
        Adaptive Enrichment / Signature Design & Inclusion criteria, \newline Target population & Error rate, Sample size & Maximizing power \\
        \toprule
        Adaptive Seamless \newline Design & Sub-trial designs & Sub-trial objectives, Error rate, Target population & Maximizing power \\
        \toprule
        \textbf{RFAN (Ours)} & Inclusion criteria, Distribution of patients across treatments & Error rate, Sample size, \newline Target population & PTMB / PTF \\
        \bottomrule[1.5pt]
    \end{tabular}
    \end{small}
    \end{center}
\end{table}

\newpage
\section{RFAN: ACQUISITION FUNCTIONS}

\begin{table}[H]
    \caption{Acquisition Functions for Clinical Trials}
    \label{appendix_tab:acquisition_functions}
    \vskip 0.1in
    \begin{center}
    \begin{small}
    \begin{tabular}{p{1.5cm} p{7cm} p{7cm}}
        \toprule[1.5pt]
        \textbf{Name} & \textbf{Formula} & \textbf{Description} \\
        \midrule[1.5pt]
        {\normalsize $\alpha_{\mu_\pi}$} & 
        $\begin{aligned}
            &\alpha_{\mu_\pi}(\tilde{\mathcal{X}}_t,\mathcal{D}_t) = \\
            &\argmax_{\{x_i\}_{i=1}^b, \{\pi_t(x_i)\}_{i=1}^b} I(Y^{\pi_{t}(x)} ; \omega |x, \pi_{t}(x), \mathcal{D}_t)
        \end{aligned}$  
        & \vspace{-4ex} Acquires patients that maximally reduce uncertainty in model parameters to predict the potential outcome, given the treatment recommended by the current policy $\pi_{t}(x)$. Patients are assigned to treatment arms according to the policy recommendation.
        \vspace{+2ex}
        \\
        \toprule
        {\normalsize $\alpha_{\mu-max}$} & 
        $\begin{aligned}
            &\alpha_{\mu-max}(\tilde{\mathcal{X}}_t,\mathcal{D}_t) = \\
            &\argmax_{\{x_i\}_{i=1}^b, {\{w_i\}_{i=1}^b}}{I(Y^w ; \omega |x, w, \mathcal{D}_t)}
        \end{aligned}$
        & \vspace{-4ex}
        Acquires and assigns patients that maximally reduce the uncertainty, under any treatment $w\in\mathcal{W}$, regardless of the current policy. 
        \vspace{+1ex}
        \\
        \toprule
        {\normalsize $\alpha_{\mu_{\pi}-max}$} & 
        $\begin{aligned}
            &\alpha_{\mu_{\pi}-max}(\tilde{\mathcal{X}}_t,\mathcal{D}_t) = \\
            &(\{x_i^*\}_{i=1}^b, \argmax_{\{w_i\}_{i=1}^b}{I(Y^w ; \omega |x^*, w, \mathcal{D}_t)}) \\
            &\text{Where } \{x_i^*\}_{i=1}^b=\argmax_{\{x_i\}_{i=1}^b} I(Y^{\pi_{t}(x)} ; \omega |x, \pi_{t}(x), \mathcal{D}_t)
        \end{aligned}$ 
        & \vspace{-7ex} 
        Acquires patients according to the current policy, as in $\alpha_{\mu_{\pi}}$. Here, the treatment assignments are according to the treatment $w\in\mathcal{W}$ that maximally reduces the uncertainty for each patient. \\
        \bottomrule
        {\normalsize $\alpha_{\mu_{\pi}-Unf}$} & 
        $\begin{aligned}
            &\alpha_{\mu_{\pi}-Unf}(\tilde{\mathcal{X}}_t,\mathcal{D}_t) = (\{x_i^*\}_{i=1}^b, \{w_i^*\}_{i=1}^b) \\
            &\text{Where } \{x_i^*\}_{i=1}^b = \argmax_{\{x_i\}_{i=1}^b} I(Y^{\pi_{t}(x)} ; \omega |x, \pi_{t}(x), \mathcal{D}_t) \\
            &\text{ and } w_i^* \sim \text{Uniform}(\mathcal{W}) 
        \end{aligned}$ 
        & \vspace{-5ex} 
        Acquires patients according to the current policy, as in $\alpha_{\mu_{\pi}}$, and assigns treatments uniformly at random. \\
        \toprule
        {\normalsize $\alpha_{\text{sign}(\tau)-\pi}$} & 
        $\begin{aligned}
            &\alpha_{\text{sign}(\tau)-\pi}(\tilde{\mathcal{X}}_t,\mathcal{D}_t) = (\{x_i^*\}_{i=1}^b, \{w_i^*\}_{i=1}^b) \\
            &\begin{aligned}
                \{x_i^*\}_{i=1}^b &= \argmax_{\{x_i\}_{i=1}^b} I(\text{sign}(Y^1 - Y^0) ; \omega |x, \mathcal{D}_t) \\
                \{w_i^*\}_{i=1}^b &= \{\pi_t(x_i^*)\}_{i=1}^b
            \end{aligned}
        \end{aligned}$ 
        & \vspace{-6ex}
        Acquires patients that maximally reduce uncertainty in model parameters to predict the treatment policy sign($\tau$)=$\mathds{1}\{\tau(x) > 0\}$. Patients are assigned to treatment arms according to the policy recommendation. In real-world clinical settings, determining a correct individual treatment assignment is often more clinically important than accurately approximating the potential outcomes themselves. $\alpha_{\text{sign}(\tau)-\pi}$ builds on this concept.
        \\
        \bottomrule[1.5pt]
    \end{tabular}
    \end{small}
    \end{center}
\end{table}

\newpage

\section{EXPERIMENTAL DETAILS}
\label{appendix:experimental_details}

Below we provide experimental details for the synthetic and semi-synthetic simulations. All experiments are run on an internal cluster consisting of several servers, including those equipped with Intel Xeon processors, AMD Opteron 6140 processors, and AMD EPYC 7702 64-Core processors. Code to implement our framework and reproduce all experiments is available at \href{https://github.com/noyomer/rfan-trial}{https://github.com/noyomer/rfan-trial} or at \href{https://github.com/Shamir-Lab/rfan-trial}{https://github.com/Shamir-Lab/rfan-trial}.

\subsection{Synthetic Simulation}
We modify the synthetic dataset presented in \cite{kallus2019interval, jesson2021causal}:
\begin{align*}
    X \sim & \quad \mathcal{N}(0,1) \\
    Y^w = & \quad (2w - 1)x + (2w - 1) - 2 \sin ((2w - 2)x) + 2(1 + 0.5x) + \epsilon
\end{align*}
Where $\epsilon \sim \mathcal{N}(0,1)$. Each random generation of the simulated data generates a pool dataset of size $10,000$ and a test set of size $2000$. During each trial, at each acquisition step, the acquired data is randomly partitioned online into training and unseen validation sets (train/validation ratio = 0.9). We define two sensitive subgroups defined as $s_1=\{x | x < -1.2\}$ and $s_2=\{x | 1.3 \geq x\}$, each reflects about $10-11\%$ of the population. In this experiment, for illustrative purposes, while patients are acquired from $D_{pool}$, no new patients can join after the trial starts. We initialize $D_{pool}$ to size $10K$. 

We investigate two trial sample sizes: $N=300$ ($T=30$) and $N=100$ ($T=10$). In both cases, the batch size, $b$, (i.e., number of patients acquired at each acquisition step) is set to 10 patients. The planning strategy for obtaining treatment policy is set to sign($\tau$)=$\mathds{1}\{\tau(x) > 0\}$. We report results over $20$ and $10$ random realizations for $N=100$ and $N=300$, respectively. The seeds are $i$ and $i+1$ ($i \in [0,...,19]$), for the training and test set, respectively (the validation set is actively collected by the acquired data of the training set).

\subsection{Warfarin Semi-Synthetic Simulation}
\paragraph{PharmGKB Data}
\label{sec:warfarin_preprocessing}

This section describes the preprocessing of the PharmGKB dataset \citep{limdi2010warfarin}. PharmGKB contains 5,700 patients who were treated with warfarin. The data for each patient include demographics (age, sex, ethnic group, weight, bmi, and smoking status), treatment reason (e.g., stroke), comorbidities (e.g., diabetes), current medications (e.g., aspirin), and genetic factors (presence of genotype variants of CYP2C9 and VKORC1). 

We included only patients with available true stable doses of warfarin and its corresponding stable observed INR (N=$4,850$). We excluded patients with no BMI available (N=$3,964$). 

All of the variables are categorical besides age, weight, and BMI. Categorical variables were encoded using \textit{one-hot encoding}, including missingness indicators (i.e., indicate which values of the variable are originally observed). A missingness indicator was added for missing age as well. Variables with small variance (standard deviation $<0.05$) were excluded. Continuous variables were standardized separately in the train and test sets of each trial. 

\paragraph{Simulation} 

We bucket the dosage into two treatment arms: low dosage ($<$ 35 mg/week) and high dosage ($\geq$ 35 mg/week) (as done in \cite{kallus2022assessing}). We consider binary outcomes that assess the therapeutics' stability and assume for simulation purposes that a warfarin dosage categorized differently from the reported stable one is considered unstable. Accordingly, for each patient, we set the outcome to $1$ ($Y=1$) if they were assigned to the arm corresponding to the patient's true optimal dose. Otherwise, the outcome is set to ($Y=0$). We define race and sex as sensitive attributes.

We investigate sample size of $N=400$ patients ($T=40$), with $t^*=20$. The batch size, $b$, (i.e., number of patients acquired at each acquisition step) is set to 10 patients. The planning strategy for obtaining treatment policy is set to sign($\tau$)=$\mathds{1}\{\tau(x) > 0\}$. We report results over $5$ random seeds. For each seed, the dataset is split into training and test subsets (train/test ratio = 0.8) using the \textit{scikit-learn} function train\_test\_split(). Then, during each trial, at each acquisition step, the acquired training data is randomly partitioned online into training and unseen validation subsets (train/validation ratio = 0.9).

\paragraph{Data Sharing} The data is publicly available at \href{https://www.pharmgkb.org/downloads}{PharmGKB repository}.

\subsection{COVID-19 Semi-Synthetic Simulation}
\label{appendix:covid_simulation}

\paragraph{SIVEP-Gripe Data}

We analyze COVID-19 hospitalization data using the SIVEP-Gripe (Sistema de Informação de Vigilância Epidemiológica da Gripe) repository \yrcite{opendatasussrag2021}, a publicly available dataset in the Brazilian Ministry of Health database. The full dataset comprised $N=99,557$ patients. The data contains clinical features such as age, sex, symptoms, comorbidities, antiviral medications, ethnic groups and regions. Patient ethnicity was classified according to five categories: Branco (White), Preto (Black), Amarelo (East Asian), Indígeno (Indigenous), and Pardo (mixed ethnicity). 

We included only patients who have tested positive for SARS-CoV-2 ($N=19,940$). We excluded patients without ethnicity recorded ($N=12,221$). we then excluded patients who were not hospitalized, resulting in a cohort of $N=11,321$ patient admissions.

\paragraph{Simulation}

We simulate an experiment that assesses the effect of receiving antiviral medications on patient survival. We consider two treatment arms: Receiving antiviral medication on the first day of hospital admission ($w=1$), or not receiving antiviral medication ($w=0$). For each patient, we set the outcome to 1 ($Y=1$) if survived at the end of hospitalization, otherwise $Y=0$. Following \cite{baqui2020ethnic}, we define region and race as sensitive attributes. 

We investigate the scenario where the treatment is potentially successful. Therefore, we undersampled non-survivors from the treated group to create a biased sample for which the survival rates of treated patients are generally higher than the non-treated patients. To estimate the counterfactual outcomes, we trained 2 \textit{XGBoost} classifiers: one classifier was trained on the treated group and the other on the control group. The trained models were then used to predict the counterfactual outcomes for each patient.

We investigate sample size of $N=400$ patients ($T=40$), with $t^*=20$. The batch size, $b$, (i.e., number of patients acquired at each acquisition step) is set to 10 patients. The planning strategy for obtaining treatment policy is set to sign($\tau$)=$\mathds{1}\{\tau(x) > 0\}$. We report results over $8$ random seeds. For each seed, the dataset is split into training and test subsets (train/test ratio = 0.8) using the \textit{scikit-learn} function train\_test\_split(). Then, during each trial, at each acquisition step, the acquired training data is randomly partitioned online into training and unseen validation subsets (train/validation ratio = 0.9).

\paragraph{Data Sharing} The data is publicly available at \href{https://opendatasus.saude.gov.br/dataset/srag-2020}{SRAG 2020 repository}.

\subsection{Statistical Testing and Early Stopping}
\label{appendix:statistical_test_and_es}

We employed the \textit{scipy.stats} Independent two-sample t-test as the hypothesis test ($\eta$), with a significance level ($\varepsilon$) of 0.05. The test was conducted on the randomized portion only, i.e., on the sample recruited in the RFAN's Randomized Stage, or on the full sample of an RCT. We set $\eta=1$ if a positive effect is detected (p-value $< \varepsilon$).

To select $t^*$ using early stopping, we conduct sequential hypothesis testing. In experiments, we use \textit{alpha spending} to responsibly adjust $\varepsilon$ values at intermediate time points, such that the overall Type 1 error rate remains at the desired level. We practically use the O'Brien-Fleming alpha spending function \citep{o1979multiple}: 
\begin{align}
    \varepsilon(t) = 2 - 2 \Phi\left(\frac{Z_{\varepsilon/2}}{\sqrt{t}}\right)
\end{align}
Where $\varepsilon$ is the overall significance level of the test, $t \in [0,1]$ is the information fraction at the interim analysis, $Z_{\varepsilon/2}$ is the upper quantile of the standard normal distribution at $\varepsilon/2$ and $\Phi$ is the normal cumulative distribution function. The information fractions for intermediate evaluation were set apriori to $[0, 0.25, 0.5, 0.75, 1]$.

\section{MODEL DETAILS}
\label{sec:appendix_model}

We use a deep-kernel GP model in our experiments. Deep-kernel GP model uses a deep feature encoder (e.g., Neural network) to transform the inputs and defines a (GP) kernel over the extracted feature representation to make predictions. The treatment variable is appended to the extracted features to form the input which is then fed into the GP.

\paragraph{Architecture}
The full model architecture is as follows:

\begin{table}[h!]
\centering
\begin{tabular}{l l}
\toprule
\textbf{Layer (type:depth-idx)} & \textbf{Param \#} \\
\midrule
DeepKernelGP & -- \\
\quad $\vert$-Sequential: 1-1 & -- \\
\quad \quad $\vert$-NeuralNetwork: 2-1 & -- \\
\quad \quad \quad $\vert$-Sequential: 3-1 & 20,400 \\
\quad \quad $\vert$-Activation: 2-2 & -- \\
\quad \quad \quad $\vert$-Sequential: 3-2 & -- \\
\quad $\vert$-VariationalGP: 1-2 & -- \\
\quad \quad $\vert$-VariationalStrategy: 2-3 & 1,515 \\
\quad \quad \quad $\vert$-CholeskyVariationalDistribution: 3-3 & 240 \\
\quad \quad $\vert$-ConstantMean: 2-4 & 1 \\
\quad \quad $\vert$-ScaleKernel: 2-5 & 1 \\
\quad \quad \quad $\vert$-RBFKernel: 3-4 & 1 \\
\quad \quad \quad $\vert$-Positive: 3-5 & -- \\
\midrule
Total params: 22,158 & \\
\bottomrule
\end{tabular}
\end{table}

\newpage
\paragraph{Hyper-parameters}
\label{section:appendix_hp}

 During each trial, the acquired data is randomly split into training and validation sets at each acquisition step. Once the trial is done, the model is tuned and retrained using the acquired unseen validation set. We use ray tune \citep{liaw2018tune} with the HyperOpt \citep{bergstra2013making} optimization algorithm for hyper-parameter tuning. The hyper-parameters search space is given in \cref{table:hp}. The hyper-parameters selected for our final model are those found to minimize the mean loss of the defined objective. 
 
\begin{table}[h!]
    \centering
    \vskip 0.1in
    \caption{Hyper-parameter search space}
    \vskip 0.1in
    \begin{tabular}{l l}
        \toprule[1pt]
        \textbf{Hyper-parameter} & \textbf{Search Space}   \\ 
        \midrule
            Kernel  & [RBF, Matern]  \\
            $\nu$ (Matern)  & [0.5, 1.5, 2.5]  \\
            inducing points  & [15, 30, 60]  \\
            hidden units  & [50, 100, 200]  \\
            network depth  & [2, 3, 4]  \\
            negative slope  & [ReLU \citep{agarap2018deep}, 0.1, ELU \citep{clevert2015fast}] \\
            dropout rate  &  [0.1, 0.2, 0.5]  \\
            spectral norm  & [None, 0.95, 1.5] \\
            batch size & [32, 64, 100, 200] \\
            learning rate  & [2e-4, 5e-4, 1e-3]  \\
        \bottomrule[1pt]
    \end{tabular}
    \label{table:hp}
\end{table}

\newpage
\section{SUPPLEMENTARY RESULTS}
\label{appendix:supp_results}

\subsection{COVID-19}

\begin{table*}[ht]
    \caption{Performance comparison on COVID-19 data (N=$400$, T=$40$, $t^*$=$20$)} 
    \label{appendix_tab:covid_performance}
    \begin{center}
    \begin{small}
    \vskip 0.1in
    \scalebox{0.8}{
    \begin{tabular}{l ccccccc}
        \toprule[1pt]
        \textbf{Design} & \textbf{Policy Val.} & \textbf{WC Policy Val.} & \textbf{\% Succ.} & \textbf{\% Policy Err.} & \textbf{$\sqrt{\epsilon_{PEHE}}$} & \textbf{PTMB (Obj. 1)} & \textbf{PTF (Obj. 2)} \\
        \midrule
        RFAN $\alpha_{\text{sign}(\tau)-\pi}$ ($t^*$=20) & 0.76 $\pm$ 0.0 & 0.47 $\pm$ 0.02 & 0.88 $\pm$ 0.12 & 7.53 $\pm$ 0.86 & 0.52 $\pm$ 0.0 & 0.73 $\pm$ 0.02 & 0.44 $\pm$ 0.02 \\ 
        RFAN $\alpha_{\mu_{\pi-max}}$ ($t^*$=20) & 0.76 $\pm$ 0.0 & 0.47 $\pm$ 0.02 & 0.88 $\pm$ 0.12 & 7.09 $\pm$ 0.76 & 0.52 $\pm$ 0.0 & 0.74 $\pm$ 0.02 & 0.44 $\pm$ 0.02 \\ 
        RFAN $\alpha_{\mu_{max}}$ ($t^*$=20) & 0.76 $\pm$ 0.0 & 0.47 $\pm$ 0.02 & 0.88 $\pm$ 0.12 & 7.09 $\pm$ 0.76 & 0.52 $\pm$ 0.0 & 0.74 $\pm$ 0.02 & 0.44 $\pm$ 0.02 \\ 
        RFAN $\alpha_{\mu_{\pi-Unf}}$ ($t^*$=20) & 0.75 $\pm$ 0.01 & 0.47 $\pm$ 0.02 & 0.88 $\pm$ 0.12 & 10.8 $\pm$ 3.88 & 0.53 $\pm$ 0.0 & 0.72 $\pm$ 0.03 & 0.44 $\pm$ 0.02 \\ 
        RFAN $\alpha_{\mu_\pi}$ ($t^*$=20) & 0.76 $\pm$ 0.0 & 0.47 $\pm$ 0.02 & 0.88 $\pm$ 0.12 & 7.45 $\pm$ 0.85 & 0.52 $\pm$ 0.0 & 0.73 $\pm$ 0.02 & 0.44 $\pm$ 0.02 \\ 
        \midrule
        Causal-BALD & 0.76 $\pm$ 0.01 & 0.46 $\pm$ 0.02 & 0.12 $\pm$ 0.12 & 7.38 $\pm$ 0.46 & 0.53 $\pm$ 0 & 0.59 $\pm$ 0.02 & 0.36 $\pm$ 0.03 \\ 
        RCT & 0.76 $\pm$ 0.0 & 0.39 $\pm$ 0.06 & 1.0 $\pm$ 0.0 & 9.97 $\pm$ 3.7 & 0.52 $\pm$ 0.01 & 0.76 $\pm$ 0.0 & 0.39 $\pm$ 0.06 \\ 
\bottomrule[1pt]
    \end{tabular}}
    \end{small}
    \end{center}
\end{table*}

\subsection{Synthetic}

We investigate the performance of RFAN with different acquisition functions, in three settings:
\begin{itemize}
    \item [(i)] Using a predefined $t^*$.
    \item [(ii)] Selection of $t^*$ using early stopping.
    \item [(iii)] Evaluation over varying $t^*$.
\end{itemize}

We run our experiments on two sample sizes of $N=300$ ($T=30$) and $N=100$ ($T=10$).

\subsubsection{N=300}

\begin{table*}[ht]
    \caption{Performance comparison on synthetic data (N=$300$, T=$30$, $t^*$=$15$)} 
    \label{appendix_tab:performance_300}
    \begin{center}
    \begin{small}
    \vskip 0.1in
    \scalebox{0.8}{
    \begin{tabular}{l ccccccc}
        \toprule[1pt]
        \textbf{Design} & \textbf{Policy Val.} & \textbf{WC Policy Val.} & \textbf{\% Succ.} & \textbf{\% Policy Err.} & \textbf{$\sqrt{\epsilon_{PEHE}}$} & \textbf{PTMB (Obj. 1)} & \textbf{PTF (Obj. 2)} \\
        \midrule
         RFAN $\alpha_{\text{sign}(\tau)-\pi}$ ($t^*$=15) & 3.17 $\pm$ 0.02 & 1.24 $\pm$ 0.04 & 1.0 $\pm$ 0.0 & 0.56 $\pm$ 0.15 & 0.77 $\pm$ 0.1 & 3.17 $\pm$ 0.02 & 1.24 $\pm$ 0.04 \\ 
        RFAN $\alpha_{\mu_{\pi-max}}$ ($t^*$=15) & 3.17 $\pm$ 0.02 & 1.23 $\pm$ 0.03 & 1.0 $\pm$ 0.0 & 0.62 $\pm$ 0.17 & 0.45 $\pm$ 0.03 & 3.17 $\pm$ 0.02 & 1.23 $\pm$ 0.03 \\ 
        RFAN $\alpha_{\mu_{max}}$ ($t^*$=15) & 3.17 $\pm$ 0.02 & 1.24 $\pm$ 0.03 & 1.0 $\pm$ 0.0 & 0.54 $\pm$ 0.17 & 0.46 $\pm$ 0.03 & 3.17 $\pm$ 0.02 & 1.24 $\pm$ 0.03 \\ 
        RFAN $\alpha_{\mu_{\pi-Unf}}$ ($t^*$=15) & 3.17 $\pm$ 0.02 & 1.22 $\pm$ 0.04 & 1.0 $\pm$ 0.0 & 1.31 $\pm$ 0.65 & 0.54 $\pm$ 0.12 & 3.17 $\pm$ 0.02 & 1.22 $\pm$ 0.04 \\ 
        RFAN $\alpha_{\mu_\pi}$ ($t^*$=15) & 3.12 $\pm$ 0.02 & 0.77 $\pm$ 0.24 & 1.0 $\pm$ 0.0 & 2.77 $\pm$ 1.08 & 1.0 $\pm$ 0.26 & 3.12 $\pm$ 0.02 & 0.77 $\pm$ 0.24 \\ 
        \midrule
        Causal-BALD & 3.17 $\pm$ 0.02 & 1.22 $\pm$ 0.03 & 0.2 $\pm$ 0.13 & 0.99 $\pm$ 0.23 & 0.54 $\pm$ 0.06 & 1.43 $\pm$ 0.3 & 1.13 $\pm$ 0.05 \\    
        RCT & 3.09 $\pm$ 0.04 & 0.55 $\pm$ 0.26 & 1.0 $\pm$ 0.0 & 4.17 $\pm$ 1.26 & 1.13 $\pm$ 0.19 & 3.09 $\pm$ 0.04 & 0.55 $\pm$ 0.26 \\  
        \bottomrule[1pt]
    \end{tabular}}
    \end{small}
    \end{center}
\end{table*}

\begin{figure}[h!]
\vskip 0.1in
\centering
\subfigure{
    \includegraphics[width=0.31\linewidth]{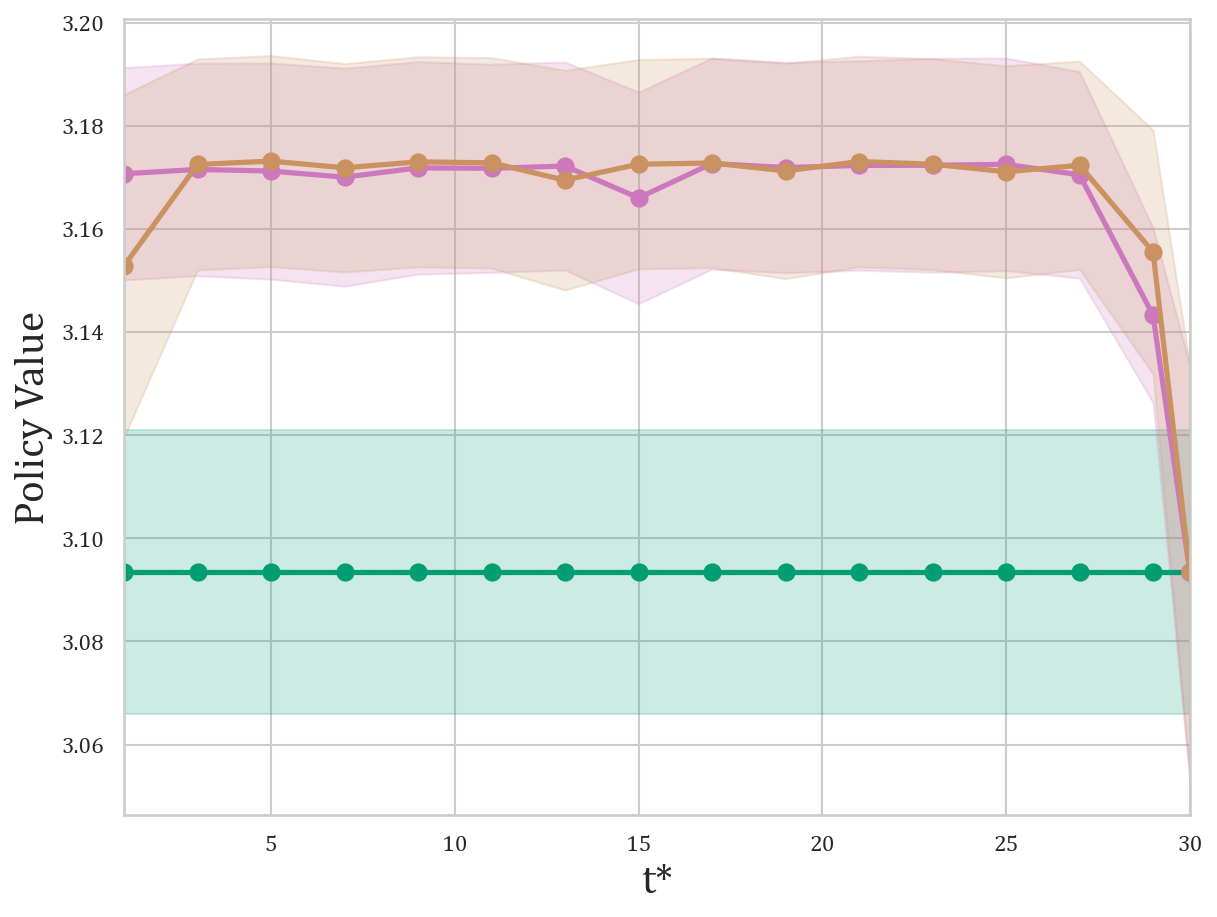}
}
\subfigure{
    \includegraphics[width=0.31\linewidth]{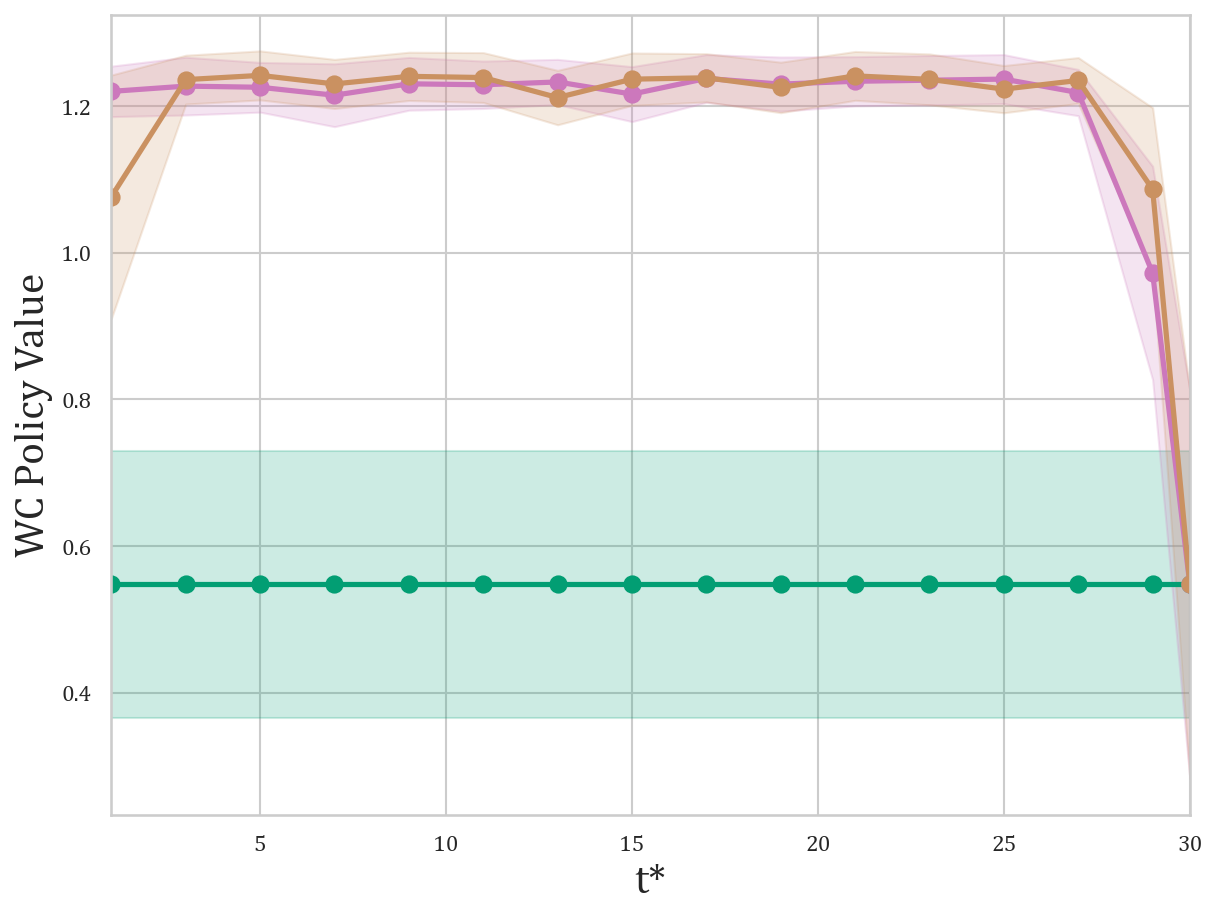}
}
\subfigure{
    \includegraphics[width=0.31\linewidth]{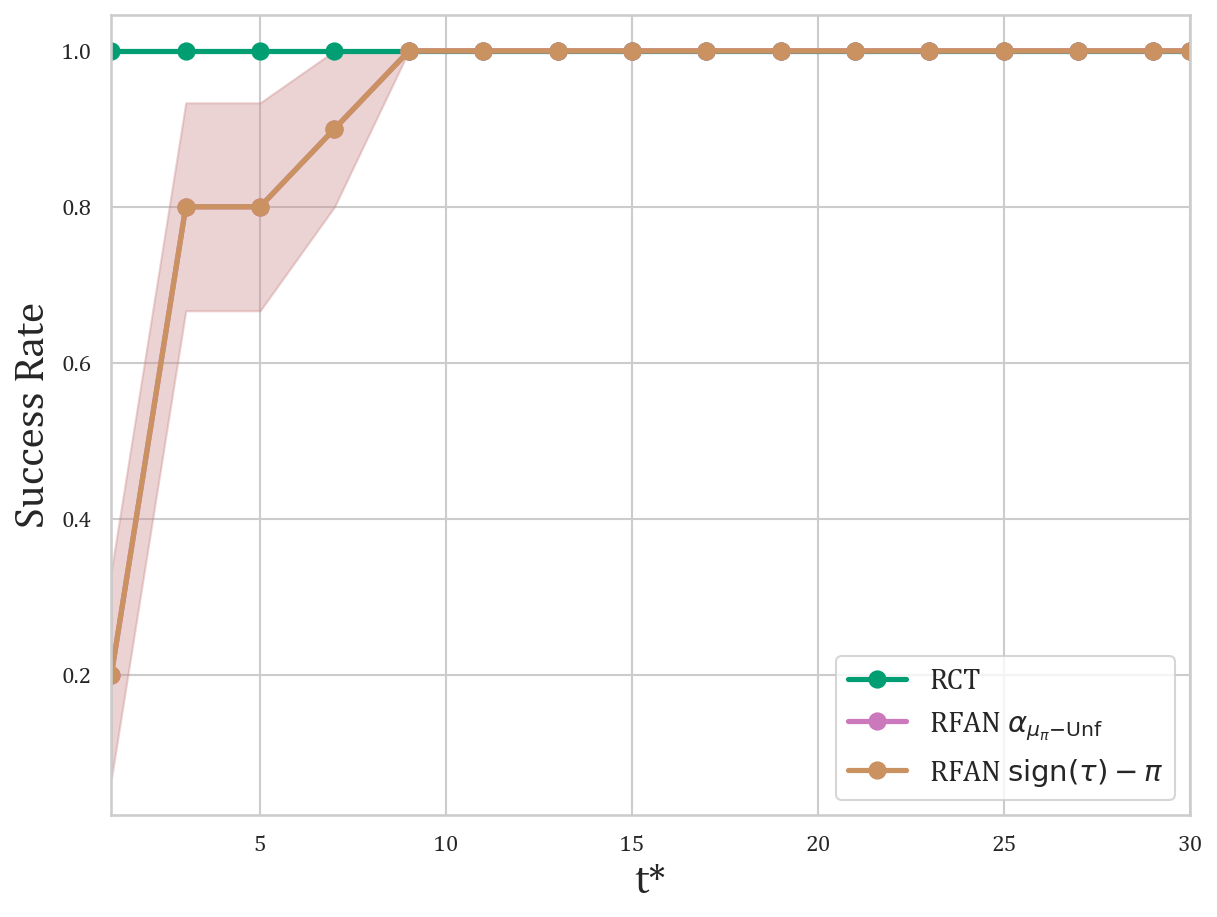}
}
\subfigure{
    \includegraphics[width=0.44\linewidth]{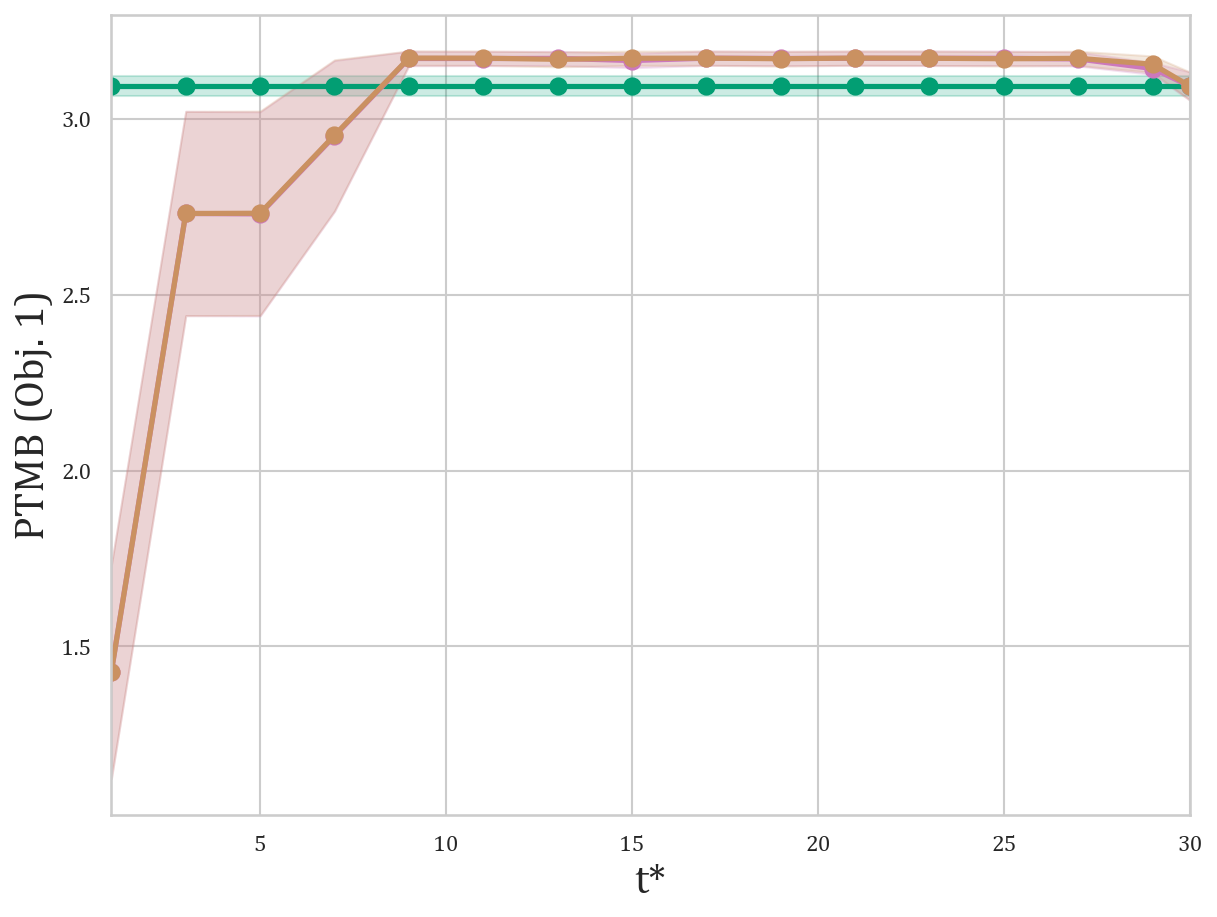}
}
\subfigure{
    \includegraphics[width=0.44\linewidth]{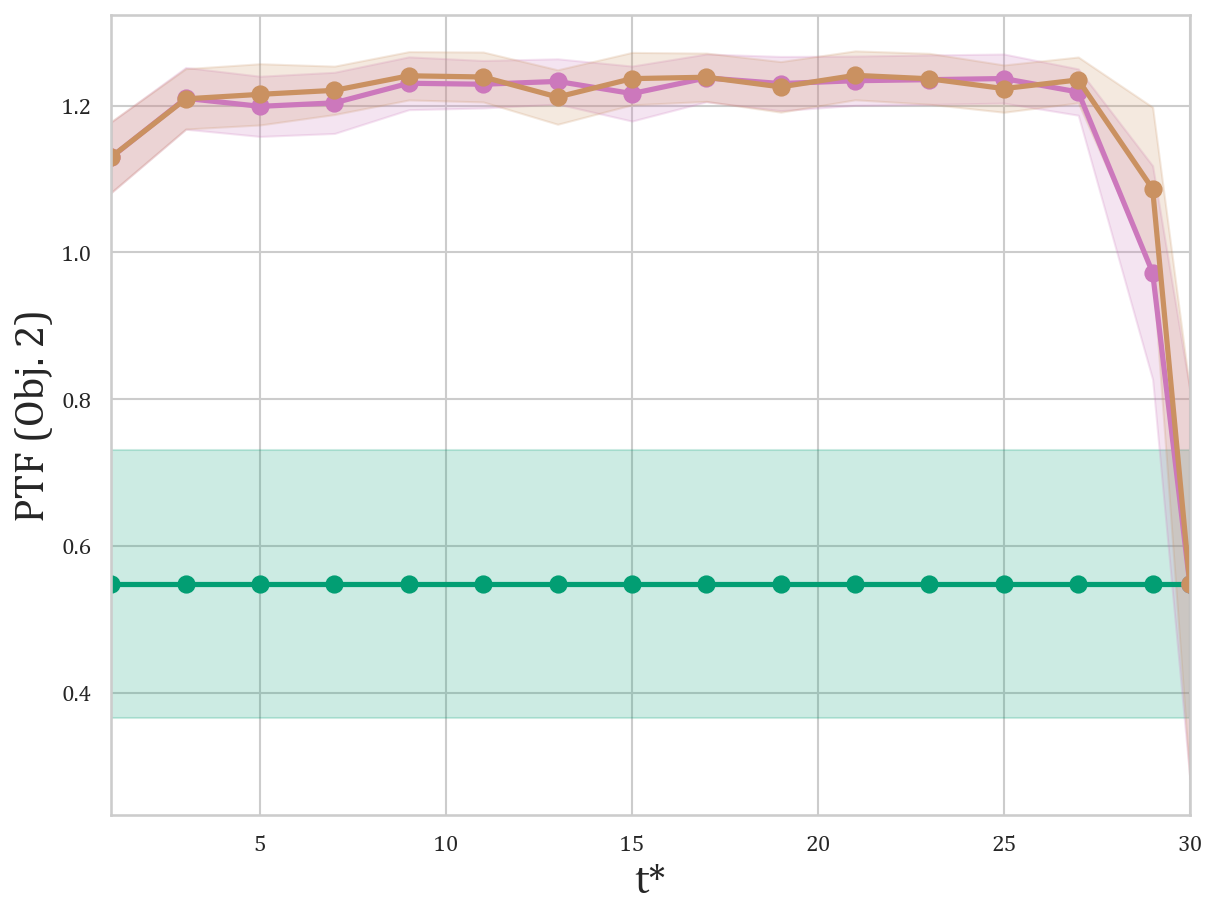}
}
\caption{Performance over varying $t*$ on synthetic data (N=$300$, T=$30$)}
\label{fig:curve_over_t_300}
\vskip -0.1in
\end{figure}

\newpage
\subsubsection{N=100}
\label{appendix:N_100}

\begin{table}[h!]
    \caption{Performance comparison on synthetic data (N=$100$, T=$10$, $t^*$=$5$)} 
    \label{tab:synthetic_performance_N100}
    \vskip 0.1in
    \begin{center}
    \begin{small}
    \scalebox{0.8}{
    \begin{tabular}{l ccccccc}
    \toprule[1pt]
         \textbf{Design} & \textbf{Policy Val.} & \textbf{WC Policy Val.} & \textbf{\% Succ.} & \textbf{\% Policy Err.} & \textbf{$\sqrt{\epsilon_{PEHE}}$} & \textbf{PTMB (Obj. 1)} & \textbf{PTF (Obj. 2)} \\ 
         \midrule
         RFAN $\alpha_{\text{sign}(\tau)-\pi}$ $(t^*=5)$ & 3.17 $\pm$ 0.01 & 1.22 $\pm$ 0.02 & 0.85 $\pm$ 0.08 & 1.29 $\pm$ 0.27 & 1.29 $\pm$ 0.08 & 2.84 $\pm$ 0.18 & 1.21 $\pm$ 0.02 \\ 
         RFAN $\alpha_{\mu_{\pi-max}}$ $(t^*=5)$ & 3.14 $\pm$ 0.03 & 1.13 $\pm$ 0.08 & 0.85 $\pm$ 0.08 & 3.16 $\pm$ 1.05 & 1.19 $\pm$ 0.09 & 2.83 $\pm$ 0.18 & 1.12 $\pm$ 0.08 \\ 
         RFAN $\alpha_{\mu_{max}}$ $(t^*=5)$ & 3.14 $\pm$ 0.03 & 1.13 $\pm$ 0.08 & 0.85 $\pm$ 0.08 & 3.2 $\pm$ 1.04 & 1.19 $\pm$ 0.09 & 2.83 $\pm$ 0.18 & 1.12 $\pm$ 0.08 \\ 
        RFAN $\alpha_{\mu_{\pi-Unf}}$ $(t^*=5)$ & 3.17 $\pm$ 0.02 & 1.23 $\pm$ 0.02 & 0.85 $\pm$ 0.08 &  1.68 $\pm$ 0.58 & 0.94 $\pm$ 0.07 & 2.84 $\pm$ 0.18 & 1.22 $\pm$ 0.02 \\ 
        RFAN $\alpha_{\mu_\pi}$ $(t^*=5)$  & 3.16 $\pm$ 0.02 & 1.19 $\pm$ 0.02 & 0.85 $\pm$ 0.08 & 2.8 $\pm$ 0.57 & 1.18 $\pm$ 0.07 & 2.83 $\pm$ 0.18 & 1.18 $\pm$ 0.02 \\  \midrule
       RFAN $\alpha_{\text{sign}\tau-\pi}$ (ES) & 3.17 $\pm$ 0.01 & 1.23 $\pm$ 0.02 & 0.95 $\pm$ 0.05 & 1.27 $\pm$ 0.23 & 1.04 $\pm$ 0.08 & 3.07 $\pm$ 0.11 & 1.23 $\pm$ 0.02 \\ 
        RFAN $\alpha_{\mu_{\pi}-max}$ (ES) & 3.17 $\pm$ 0.01 & 1.2 $\pm$ 0.02 & 0.95 $\pm$ 0.05 & 2.24 $\pm$ 0.32 & 1.04 $\pm$ 0.08 & 3.06 $\pm$ 0.11 & 1.2 $\pm$ 0.02 \\ 
        RFAN $\alpha_{\mu-max}$ (ES) & 3.17 $\pm$ 0.01 & 1.2 $\pm$ 0.03 & 0.95 $\pm$ 0.05 & 2.17 $\pm$ 0.3 & 1.04 $\pm$ 0.08 & 3.06 $\pm$ 0.11 & 1.2 $\pm$ 0.03 \\ 
        RFAN $\alpha_{\mu_{\pi}-Unf}$ (ES) & 3.17 $\pm$ 0.01 & 1.2 $\pm$ 0.03 & 0.95 $\pm$ 0.05 & 1.99 $\pm$ 0.41 & 1.09 $\pm$ 0.08 & 3.06 $\pm$ 0.11 & 1.2 $\pm$ 0.03 \\ 
        RFAN $\alpha_{\mu_pi}$ (ES) & 3.16 $\pm$ 0.02 & 1.11 $\pm$ 0.08 & 0.95 $\pm$ 0.05 & 2.18 $\pm$ 0.43 & 1.16 $\pm$ 0.09 & 3.05 $\pm$ 0.11 & 1.11 $\pm$ 0.08 \\ 
        \midrule
        Causal-BALD & 3.0 $\pm$ 0.08 & 1.2 $\pm$ 0.03 & 0.3 $\pm$ 0.11 & 9.12 $\pm$ 3.15 & 1.33 $\pm$ 0.11 & 1.65 $\pm$ 0.23 & 1.16 $\pm$ 0.03  \\ 
        RCT & 3.07 $\pm$ 0.02 & 0.36 $\pm$ 0.17 & 1.0 $\pm$ 0.0 & 5.35 $\pm$ 0.8 & 1.46 $\pm$ 0.13 & 3.07 $\pm$ 0.02 & 0.36 $\pm$ 0.17 \\ 
        \bottomrule[1pt]
    \end{tabular}}
     \end{small}
    \end{center}
\end{table}

\begin{figure}[h!]
\vskip 0.1in
\centering
\subfigure{
    \includegraphics[width=0.31\linewidth]{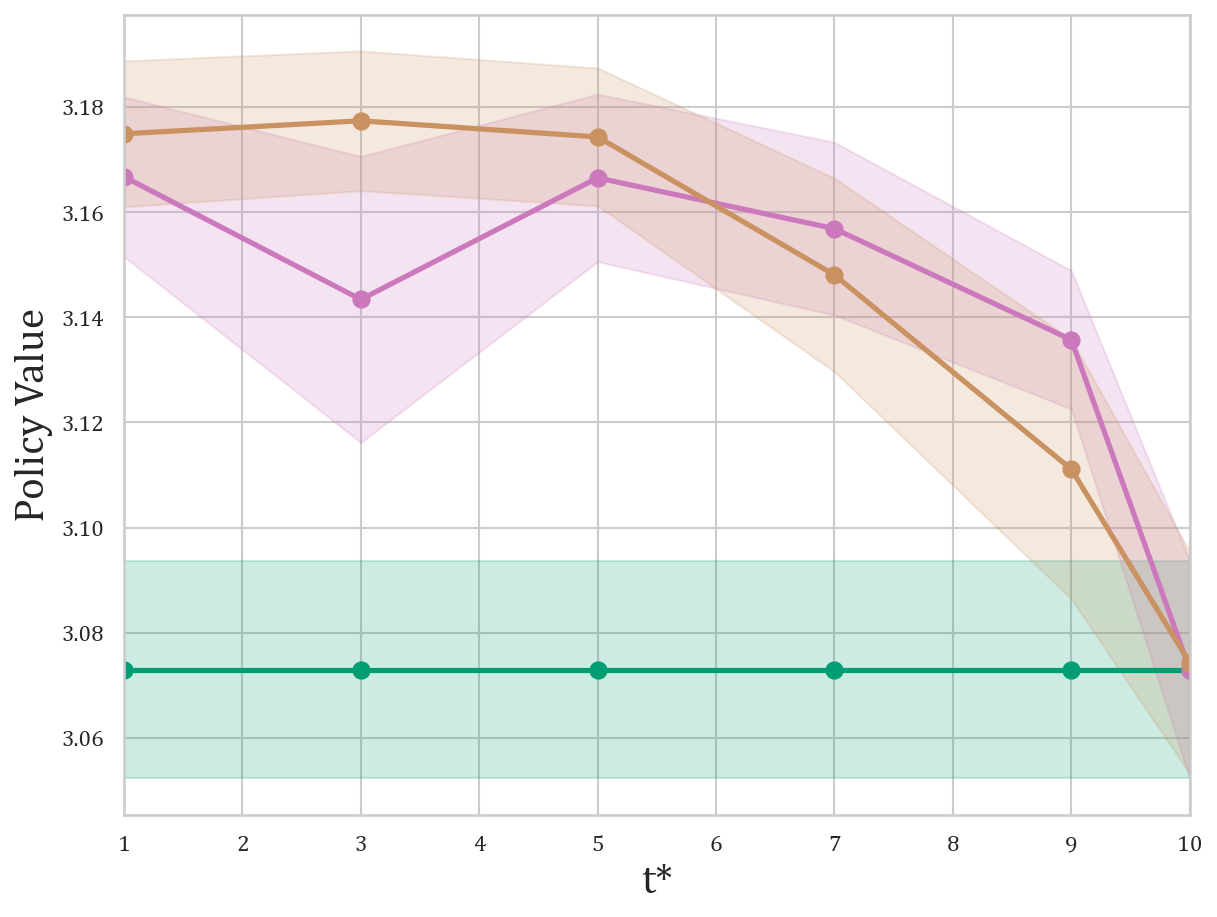}
}
\subfigure{
    \includegraphics[width=0.31\linewidth]{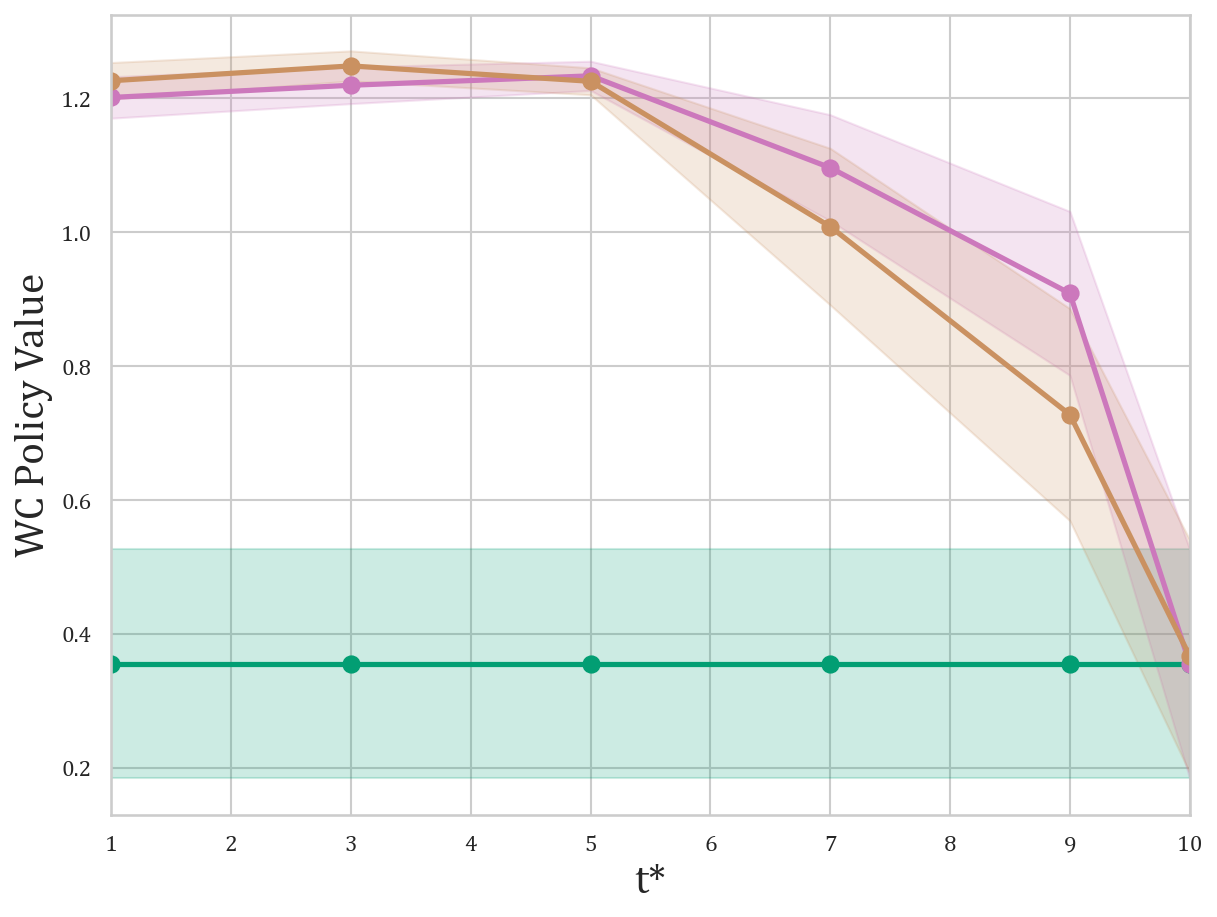}
}
\subfigure{
    \includegraphics[width=0.31\linewidth]{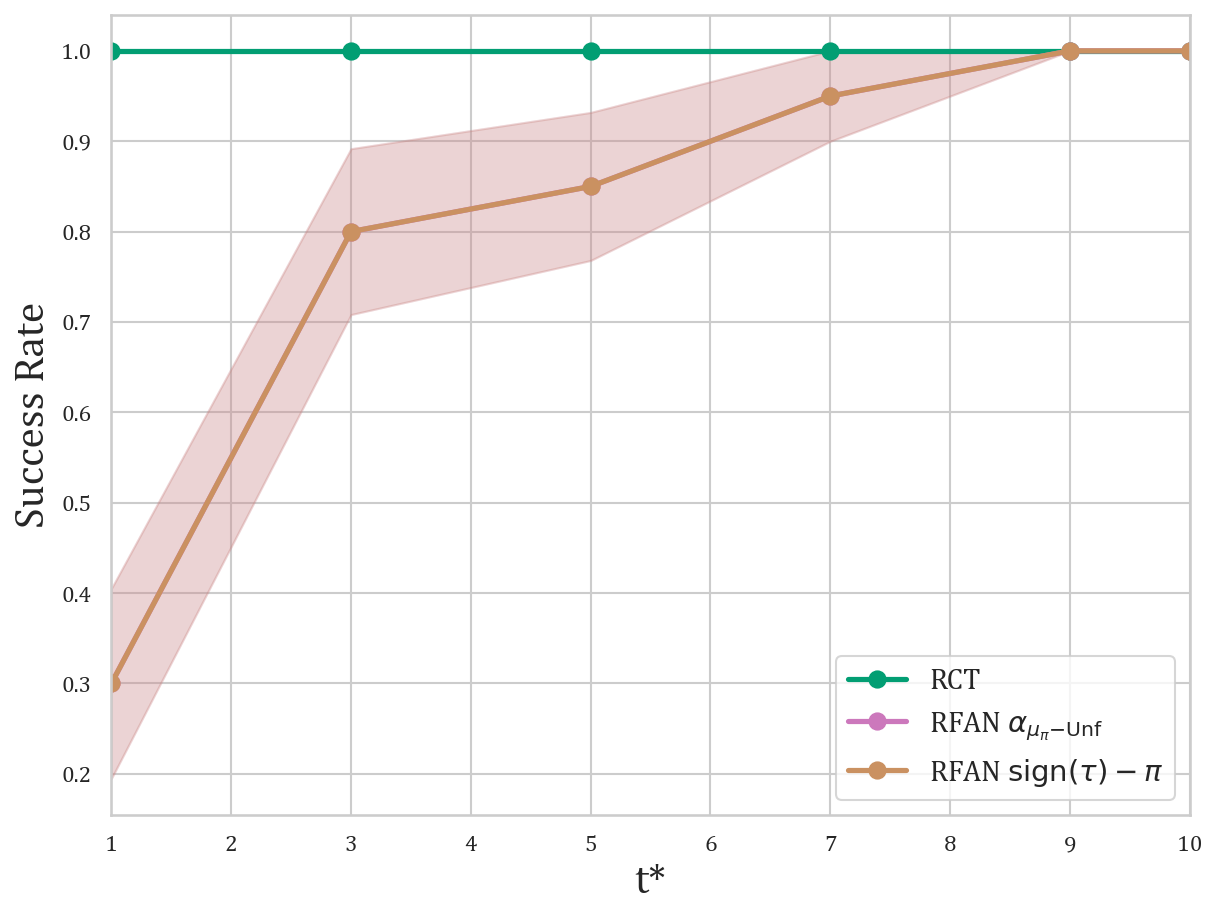}
}
\subfigure{
    \includegraphics[width=0.44\linewidth]{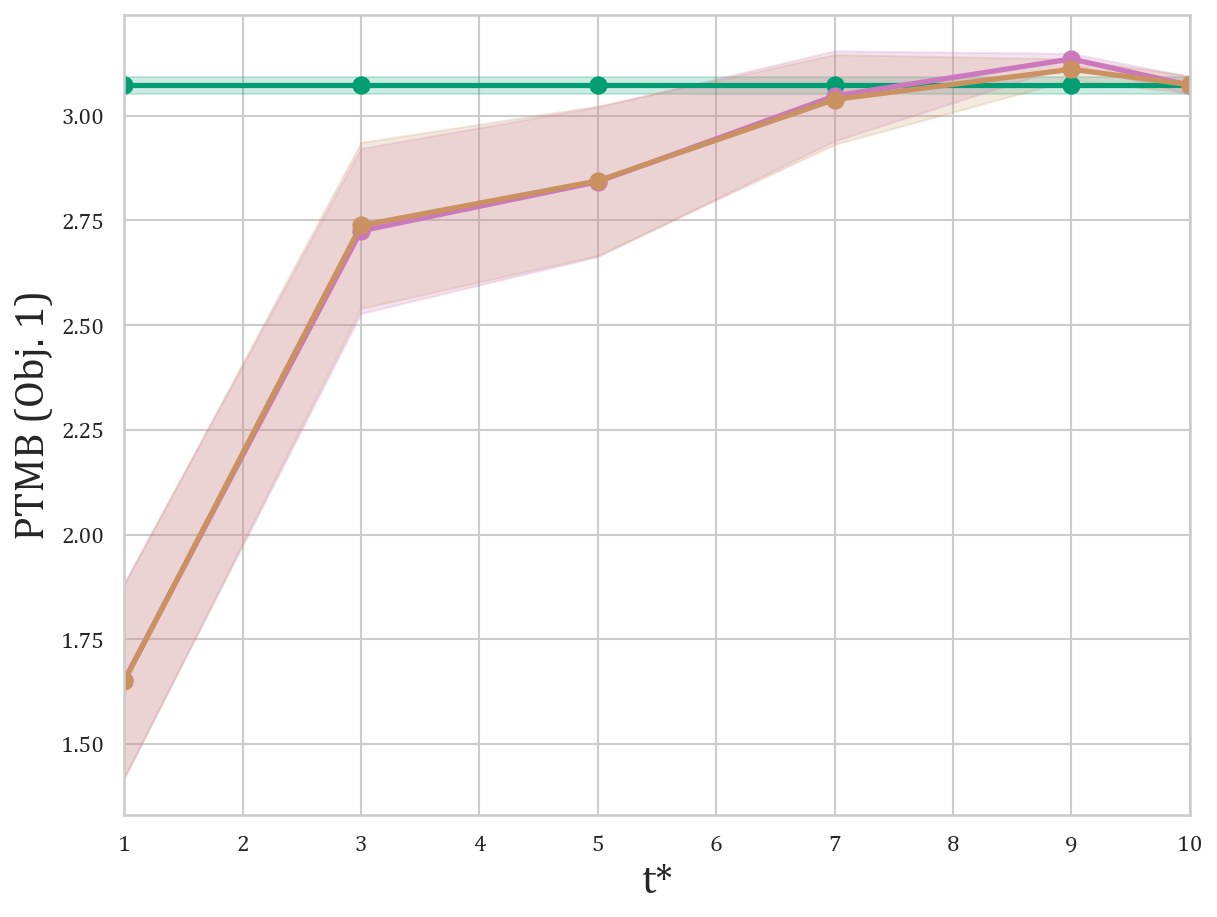}
}
\subfigure{
    \includegraphics[width=0.44\linewidth]{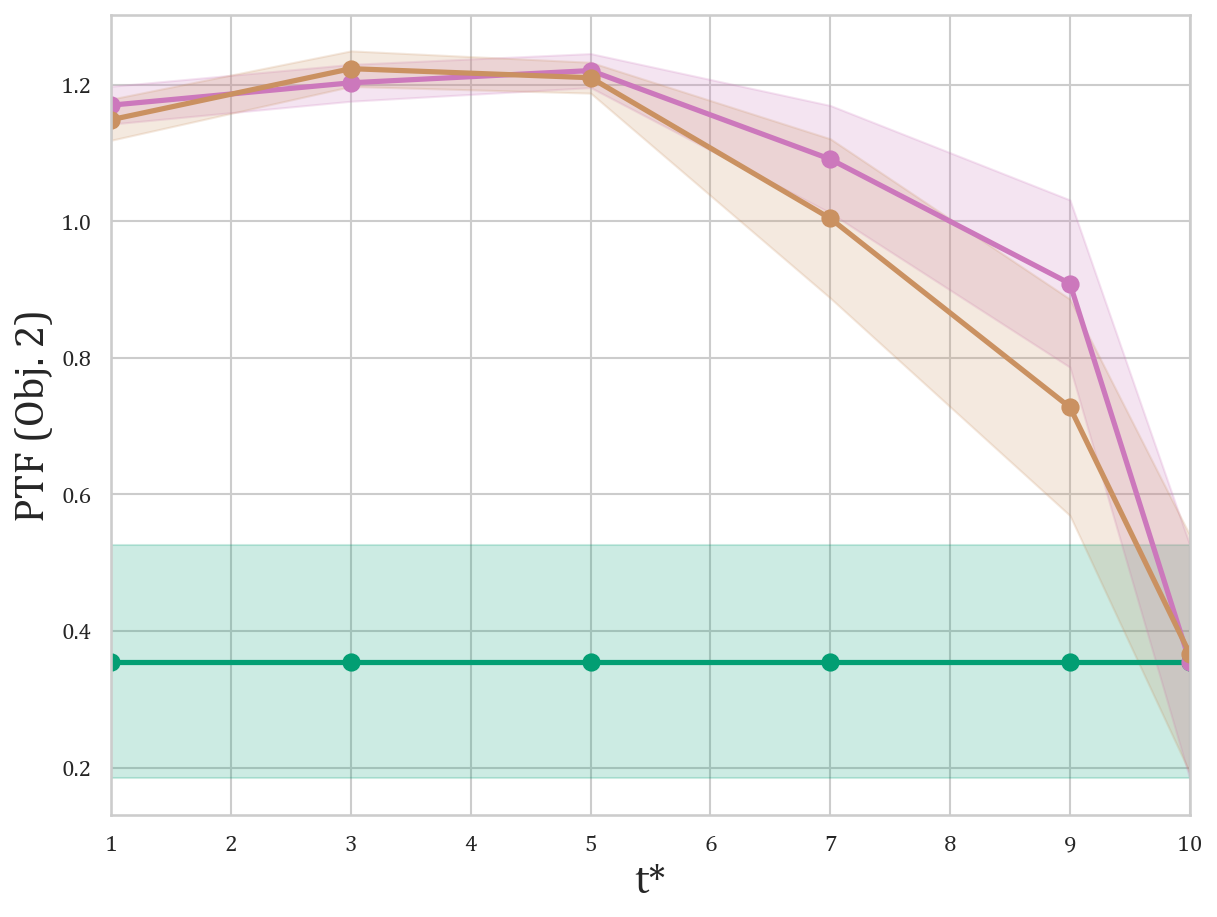}
}
\caption{Performance over varying $t*$ on synthetic data (N=$100$, T=$10$)}
\label{fig:curve_over_t_100}
\vskip -0.1in
\end{figure}

\newpage
\section{ETHICAL CONSIDERATIONS} 
\label{appendix:ethical_considerations}

The ethical justification of clinical trials relies on the principle of clinical equipoise. Clinical equipoise establishes two conditions to be met: (i) There must be uncertainty among domain experts regarding the effectiveness of a treatment, and (ii) the trial should be designed to resolve this uncertainty \citep{miller2007clinical, freedman2017equipoise}. As demonstrated in our proposed objectives, our trial is designed to learn and optimize how to treat patients once the treatment is widely deployed. In that sense, we believe that our design is more suited to the clinical equipoise, by reducing uncertainty about the treatment in practice.

The choice of employing adaptive trial designs as opposed to conventional non-adaptive trials is highly situational \citep{palmer1999ethics, fillion2019clinical}. In many scenarios, conventional trial designs are still preferable, e.g., when there are long delays between patient enrollment and the observation of outcomes, or if a homogeneous response of the target population is expected. However, there are various cases in which RFAN is expected to be beneficial. We believe that employing RFAN would be of most value when there is high uncertainty regarding the effectiveness of the new treatment and when the treatment responses are expected to be highly heterogeneous among different subpopulations. We note that it should be only considered when the pool of patients eligible for enrollment shares the same characteristics as the target population. In such cases, we believe that our design is not only ethically permissible but also has the potential to improve patient care with a confirmatory guarantee. 

Validating RFAN in real-world settings would essentially entail running a new trial. Conducting such a trial is beyond our scope, as a method development paper (discussed further in \cref{appendix:evaluation}). We therefore adhere to the standard approach of evaluating adaptive designs using simulated data (e.g., \cite{atan2019sequential, huyuk2022make, curth2023adaptive,friede2012conditional, magnusson2013group, stallard2014adaptive, henning2015closed, rosenblum2016group}.

\section{REGULATORY CONSIDERATIONS} 
\label{appendix:regulatory_considerations}

Regulatory approval is a crucial barrier for any clinical trial design. Our framework, RFAN, is purposely designed to narrow the gap between the regulatory requirements (i.e., trial success) and real-world treatment performance, without requiring a complete overhaul of the regulatory procedures currently in place. The first stage of RFAN follows a conventional randomized design, incorporating only well-established regulatory-compliant components, such as early stopping. This ensures that the initial phase of the trial aligns with existing standards, making approval relatively straightforward. The adaptive stage is designed to enhance fairness and real-world policy impact, compared to current practice, without compromising regulatory rigor. To assess RFAN's applicability and feasibility, we consulted with domain experts in clinical pharmacology and drug development. Their feedback affirmed the practical relevance and validity of our framework, ensuring that it addresses pressing concerns in current clinical trial designs. While regulatory bodies remain a key audience, we also emphasize the potential value of our methodology for other key healthcare stakeholders such as payers, pharmaceutical companies, caregivers, and healthcare providers.


\end{document}